\pdfoutput=1

\documentclass[11pt]{article}

\usepackage[final]{acl}

\usepackage{times}
\usepackage{latexsym}
\usepackage{times}
\usepackage{latexsym}
\usepackage{graphicx}
\usepackage{fdsymbol}
\usepackage{adjustbox}
\usepackage{soul}
\usepackage{footmisc}
\usepackage{enumitem}
\usepackage{xcolor}
\usepackage{caption}
\usepackage{tablefootnote}
\usepackage{multirow}

\usepackage[T1]{fontenc}

\usepackage[utf8]{inputenc}

\usepackage{microtype}

\usepackage{inconsolata}

\usepackage{graphicx}

\usepackage{color-edits}
\addauthor{gn}{magenta}
\addauthor{iw}{olive}
\addauthor{sp}{purple}
\addauthor{vv}{teal}
\addauthor{sw}{blue}

\newcommand{\Hquad}{\hspace{0.3em}} 
\newcommand\blfootnote[1]{%
  \begingroup
  \renewcommand\thefootnote{}\footnote{#1}%
  \addtocounter{footnote}{-1}%
  \endgroup
}

\title{Synthetic Multimodal Question Generation}

\author{Ian Wu$^*$$^\diamondsuit$ \Hquad Sravan Jayanthi$^*$$^\spadesuit$$^\dagger$ \Hquad Vijay Viswanathan$^\clubsuit$ \Hquad Simon Rosenberg$^\diamondsuit$ \AND \vspace{6em} Sina Pakazad$^\diamondsuit$ \Hquad  Tongshuang Wu$^\clubsuit$ \Hquad Graham Neubig$^\clubsuit$ \AND \\\vspace{-8.2em} \\ $^\diamondsuit$C3 AI \qquad $^\spadesuit$Connectly AI \qquad $^\clubsuit$Carnegie Mellon University \vspace{0.5em} \\ \texttt{\{ian.wu, simon.rosenberg, sina.pakazad\}@c3.ai}, \texttt{sravan@connectly.ai} \\ \texttt{\{sherryw, gneubig\}@cs.cmu.edu, vijayv@andrew.cmu.edu}} 

\begin{document}
{\makeatletter\acl@finalcopytrue
  \maketitle
}

\begin{abstract}
Multimodal Retrieval Augmented Generation (MMRAG) is a powerful approach to question-answering over multimodal documents. A key challenge with evaluating MMRAG is the paucity of high-quality datasets matching the question styles and modalities of interest. In light of this, we propose \textbf{SMMQG}, a synthetic data generation framework. SMMQG leverages interplay between a retriever, large language model (LLM) and large multimodal model (LMM) to generate question and answer pairs directly from multimodal documents, with the questions conforming to specified styles and modalities. We use SMMQG to generate an MMRAG dataset of 1024 questions over Wikipedia documents and evaluate state-of-the-art models using it, revealing insights into model performance that are attainable only through style- and modality-specific evaluation data. Next, we measure the quality of data produced by SMMQG via a human study. We find that the quality of SMMQG-generated synthetic data is on par with the quality of the crowdsourced benchmark MMQA and that downstream evaluation results using both datasets strongly concur.
\end{abstract}

\section{Introduction}\label{sec:introduction}
\blfootnote{$^\dagger$Work done while at C3 AI.}

\begin{figure}[htb]
    \centering
    \includegraphics[width=0.45\textwidth]{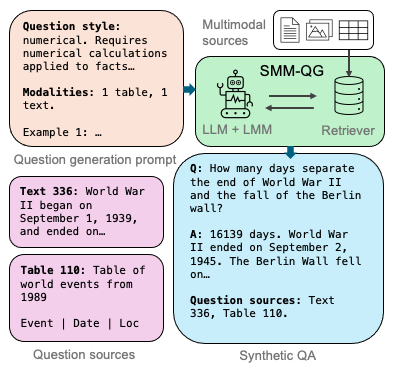}
    \caption{\textbf{An overview of SMMQG.} Given user-provided question style and modality requirements, SMMQG selects question sources and produces questions and answers. The questions are grounded in the selected question sources, and adhere to the question and modality requirements.}
    \label{fig:intro_diagram}
\end{figure}

Following the increased adoption of Retrieval Augmented Generation (RAG) \cite{lewis2021retrievalaugmented} for text-based question-answering (QA), there has been much interest in extending RAG to the multimodal setting \cite{chen2022murag, chang2022webqa, lin2022retrieval, yasunaga2023retrievalaugmented}. In multimodal RAG (MMRAG), QA is performed by a large language model (LLM) or large multimodal model (LMM) grounded in sources that span modalities such as text, tables and images. As with RAG, MMRAG has the potential to increase answer quality and transparency when compared with closed-book QA, where models directly answer questions without access to external knowledge.

A major challenge encountered when implementing MMRAG systems is evaluation, which is typically done using fixed benchmark datasets. These datasets consist of (\textit{source(s), question, answer}) tuples that enable separate evaluation of the retriever and the QA model. Some representative datasets include MMQA \cite{talmor2021multimodalqa}, MMMU \cite{yue2023mmmu} and WebQA \cite{chang2022webqa}.

The problem with this approach is that we cannot tailor the evaluation questions to our desired specifications. We identify two aspects of the evaluation questions we may wish to control. The first aspect is the \textit{question style}. The style of a question determines the reasoning abilities required to answer it, and certain models perform better on certain question styles than others. Examples of question styles include mathematical \cite{hendrycks2021measuring, yu2023metamath}, multi-hop \cite{yang2018hotpotqa} and extractive \cite{rajpurkar2016squad} question styles. The modality of a question refers to the modality or modalities (e.g. text, table, image) of the sources required to answer it, and both retrieval \cite{wei2023uniir} and QA \cite{jia2021scaling, chen-etal-2020-hybridqa} performance depends on the modalities of their inputs.

Given the sensitivity of MMRAG performance to question styles and modalities, we want our evaluation questions to match the styles and modalities of the questions our system is likely to encounter. Such a benchmark may not exist, making it difficult to perform the comprehensive evaluations needed to reveal important model deficiencies.

To address this issue, we introduce a method for \textbf{Synthetic Multimodal Question Generation} we call \textbf{SMMQG}. SMMQG is a synthetic data generation framework that leverages interplay between a retriever, an LLM and an LMM with in-context learning \cite{brown2020language} to generate multimodal questions and answers based directly on input documents. Crucially, SMMQG enables fine-grained control over the styles and modalities of questions, and is capable of producing both unimodal and cross-modal questions. In order to demonstrate the utility of SMMQG, we first use it to create a dataset of 1024 questions and answers of various styles and modalities from Wikipedia documents. We then use this dataset to unearth novel style- and modality-specific insights into the MMRAG performance of various models. Such insights would be difficult to obtain without style- and modality-specific evaluation data, which SMMQG can produce in an automatic and scalable fashion.

One concern with synthetic data generation is that the resulting data is of low quality. To assuage this concern and verify that SMMQG produces high quality data, we conduct a human study to measure the quality of our synthetic dataset. We find that our dataset's quality is on par with or better than that of popular crowdsourced benchmark MMQA \cite{talmor2021multimodalqa} when measured along five different metrics. We also show that downstream evaluation results obtained using our SMMQG dataset display strong concurrence \cite{liu2023question} with those obtained using MMQA, demonstrating that our synthetic dataset can be used in place of MMQA for model selection.

\section{Problem Setting}\label{sec:problem_setting}
\subsection{Multimodal Sources}\label{sec:multimodal_sources}

We define the multimodal sources $S$ to include the text passages, tables, and images extracted from one or more specified documents. Multimodal sources are parsed from a document as a pre-processing step.\footnote{We are agnostic to the parsing strategy \cite{zhao2023retrieving}, so long as contents of different modalities are parsed as separate sources.} Text sources consist of chunked text passages and table sources consist of pipe-separated strings that are prepended with titles. Image sources consist of an image, its caption and an image \textit{verbalisation}. The image verbalisation is a text description of the image generated using an image-captioning model or LMM. Verbalisations are useful because they allow text-based models to search and reason over images, as seen in \citet{liu2023universal}. See Appendix \ref{sec:appendix_dataset_examples} for examples of sources and Appendix \ref{sec:appendix_verbalisation} for details of our image verbalisation strategy.

\subsection{Formulation}\label{sec:overview}
\begin{figure*}
    \centering
    \includegraphics[width=1.0\textwidth]{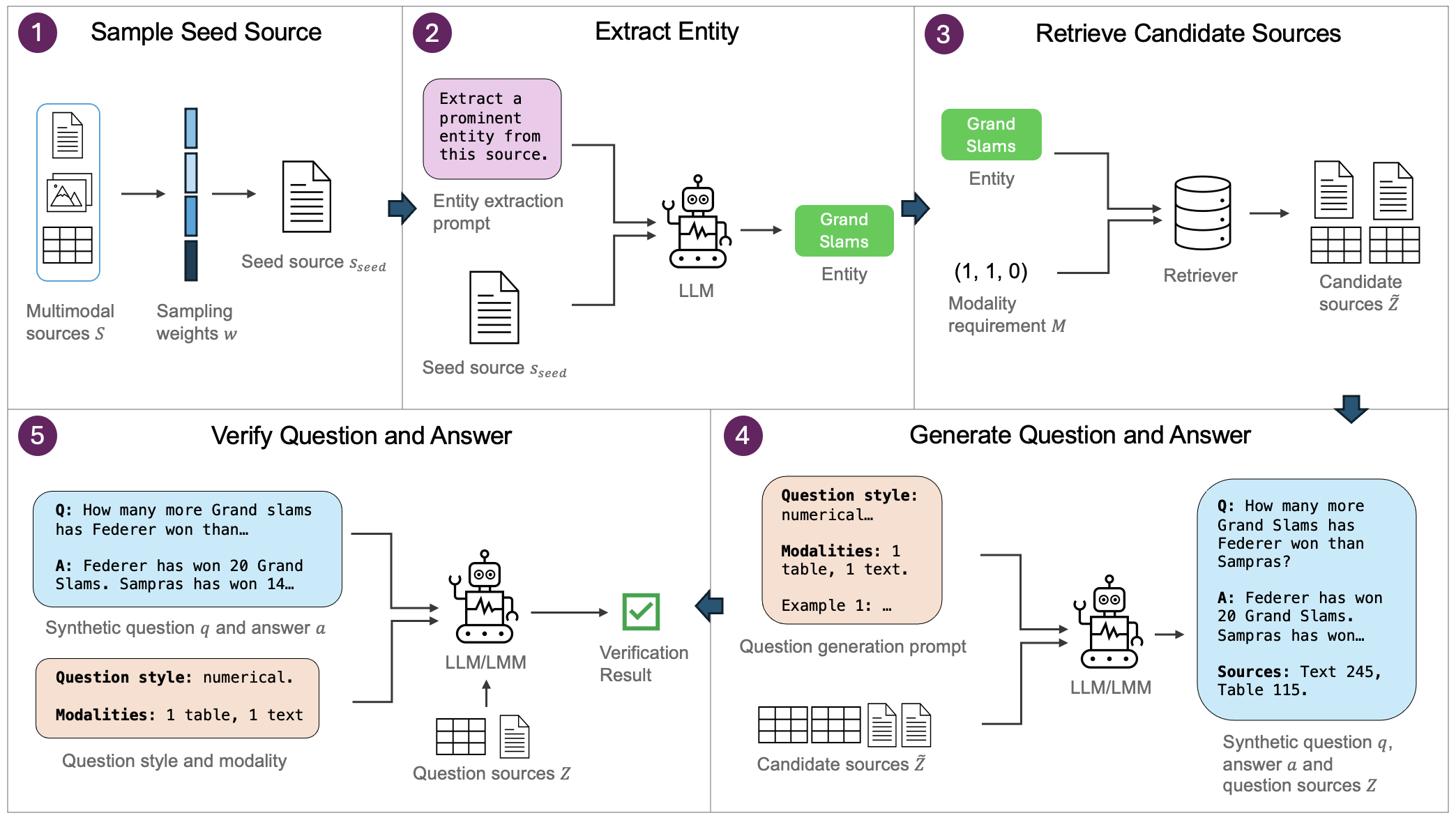}
    \caption{\textbf{SMMQG consists of five steps.} (1) A seed source is sampled from the sources $S$. (2) An entity is extracted from the seed source. (3) Candidate sources are retrieved from $S$ using the entity from step 2 as the query. (4) The question generation model chooses the question sources from amongst the candidate sources, and uses these to produce the question and answer. (5) The model is asked to verify that the generated question adheres to the desired question style and modalities, and that the generated answer correctly answers the question.}
    \label{fig:main_diagram}
\end{figure*}

SMMQG takes three inputs. These are (1) multimodal sources $S$, which serve to provide the context for question generation (2) question style $v$, which is a description of the question style along with examples, and (3) modality requirements $M$. This is a 3-tuple of integers $M = (m_{\mathrm{text}}, m_{\mathrm{table}}, m_{\mathrm{image}})$. $M$ is used to indicate the modalities of the generated questions: $M = (2, 1, 0)$, for example, indicates that our generated question should be a cross-modal text-table question with two text and one table sources.

SMMQG jointly produces three outputs. These are (1) the synthetic question $q$, with style dependent on $v$ (2) a long-form answer $a$ to the question and (3) 
references to the question sources $Z$, where $z_i \in S$. $q$ is only answerable using information derived from \textit{every} source in $Z$. The modalities of the question sources must match $M$: if $M = (1, 1, 0)$, for example, then $|Z| = 2$ and $z_1$ and $z_2$ must be text and table modalities in some order. 

\section{Method}\label{sec:method}
\subsection{SMMQG}
SMMQG is composed of five steps, as illustrated in Figure \ref{fig:main_diagram}. The output of the first three steps is a set of \textit{candidate sources} $\tilde{Z}$, where $\tilde{z}_i \in S$. These consist of semantically-related sources connected by an entity. The fourth step is responsible for generating $q$ and $a$ and for choosing $Z \subseteq \tilde{Z}$. Thematic unity is maintained by the fact that all candidate sources are semantically related, which enables the generation of meaningful multi-source and potentially cross-modal questions. The fifth step is responsible for validating $q$ and $a$.

\textbf{Step 1: Sample Seed Source}\quad The goal of this step is to locate a seed source $s_{\mathrm{seed}} \in S$. The most straightforward way to select $s_{\mathrm{seed}}$ is to choose one by uniformly sampling over $S$. However, we find that many sources are outliers that are unrelated to other sources and do not reflect the main topics found in the documents. Building coherent multi-source questions from such sources is difficult, as there are often no related candidate sources to choose from.

To correct for this, we introduce weights $w_i$. The probability of sampling $s_i$ as $s_{\mathrm{seed}}$ is
\begin{align*}
    p_{s_i} = \frac{\exp({-\beta w_i})}{\sum_j \exp{(-\beta w_j)}}
\end{align*}
where $\beta$ is a temperature parameter, which we set to 0.1 based on manual inspection of resulting outputs. The weights $w_i$ are the average cosine-distance of the $k_{\mathrm{seed}}$-nearest neighbours of $s_i$ in embedding space. Given some embedding model $E$,
\begin{align*}
    w_i = \frac{1}{k_{\mathrm{seed}}}\sum_{s_j \in k_{\mathrm{seed}}\mathrm{nn}(s_i)} \mathrm{dist}(E(s_i), E(s_j))
\end{align*}
We use the E5-Large embedding model \cite{wang2024text} in our experiments, with $k_{\mathrm{seed}} = 5$.

\textbf{Step 2: Extract Entity}\quad In this step, we use GPT-4-Turbo \cite{openai2024gpt4} to extract a prominent entity (e.g. ``tennis'', ``Japan'', ``machine learning'') from the seed source via three-shot prompting. For image sources, we provide the LLM with the image verbalisation and the image caption, which we find captures image entities sufficiently well. We also find that using a high temperature of 1.0 improves the diversity of extracted entities, which in turn improves the diversity of generated questions. See Appendix \ref{sec:appendix_smmqg_prompts} for our entity extraction prompt.

\textbf{Step 3: Retrieve Candidate Sources}\quad In this step, we retrieve candidate sources $\tilde{Z}$ using the E5-Large retriever, with the extracted entity from Step~2 as the query. The candidate sources are therefore all semantically related through the entity. 

We also define an integer $k_{\mathrm{modality}}$. For every modality $i \in \{\textrm{text}, \textrm{table}, \textrm{image}\}$, we retrieve the top $m_i k_{\mathrm{modality}}$ candidate sources of that modality. For example, given $M = (1, 2, 0)$ we retrieve $k_{\mathrm{modality}}$ text and $2k_{\mathrm{modality}} $ table sources respectively. We set $k_{\mathrm{modality}} = 2$, as we find that providing a choice of candidate sources improves the quality of generated questions. For retrieving images, we rely on image verbalisations. We find that this text-only retrieval approach improves over the use of multimodal retrieval with CLIP \cite{radford2021learning} embeddings, likely due to the prevalence of long text and table sources.

\textbf{Step 4: Question Generation} We now pass $\tilde{Z}$ to the question-generation model, which is an LLM or LMM, depending on whether there are image candidate sources present. We use GPT-4-Turbo \cite{openai2024gpt4} as both in our experiments. In addition to $\tilde{Z}$, the question-generation model is also given the task instruction, the question style and its description $v$, the modality requirements $M$, and three style-specific few-shot examples. The task instruction asks the model to choose question sources $Z$ from $\tilde{Z}$, adhering to $M$, and to then generate $q$ and $a$ using $Z$, following $v$. The model is also asked to produce references to $Z$ by outputting their source IDs.\footnote{Candidate sources that are not chosen could be used as \textit{distractors} \cite{talmor2021multimodalqa}, as they are semantically similar to the chosen sources.} We instruct the model to refuse requests if it does not believe that a coherent question adhering to $v$ and $M$ can be generated. See Appendix \ref{sec:appendix_smmqg_prompts} for our question-generation prompts.

\textbf{Step 5: Question Verification} We perform three checks and reject samples that fail any of these. Firstly, we cross-reference the modalities of the chosen question sources with $M$ and reject samples where the source modalities do not match the requirements. For example, if $M = (0, 1, 1)$, we check that the modalities of $Z$ are exactly 1 table and 1 image. Secondly, we pass $q$ to an LLM and prompt it to verify that the question adheres to the specified question style. Lastly, we pass $q$, $a$ and $Z$ to an LLM or LMM and prompt it to verify that (a) $a$ is correct given $q$ and $Z$ and (b) \textit{every} source in $Z$ is required to answer $q$. We perform both the second and third steps of QA verification with a single call to GPT-4-Turbo. See Appendix \ref{sec:appendix_qa_verification} for our QA verification prompts.

\subsection{Multi-hop QA Generation}
Although Step~4 of SMMQG can be used to generate multi-hop questions, we propose an alternative version that produces higher quality multi-hop questions more consistently. We start by generating two \textit{intermediate questions} and their respective answers and question sources: for the first intermediate question, we prompt the model to generate a \textit{question about the entity} extracted in Step~2, given some subset of $\tilde{Z}$. For the second question, we prompt the model to generate a question \textit{where the answer is the same entity}, given the remaining $\tilde{Z}$. When we generate cross-modal multi-hop questions, $\tilde{Z}$ are split by modality. When we generate unimodal multi-hop questions, $\tilde{Z}$ are split randomly.

Next, we prompt the model to \textit{combine} the intermediate questions and answers to form a multi-hop question and answer. The multi-hop question sources are the union of the question sources chosen in the intermediate steps. See Appendix \ref{sec:appendix_multi_hop_qg} for examples of the question combination prompts and a diagram illustrating multi-hop question generation.

\begin{table*}[]
    \small
    \begin{tabular}{p{0.10\linewidth} | p{0.27\linewidth} p{0.29\linewidth} p{0.17\linewidth} c}
     \hline
     \textbf{Style} & \textbf{Description} & \textbf{Example} & \textbf{Modalities} & \textbf{\#} \\\hline
     Information Extraction & Requires extracting and returning information from a single source. & What was the target age group for the NBC Kids programming block? & Text, Table, Image & 158 \\ \hline Compare Contrast & Requires comparing and contrasting two closely related entities or topics. & Compare and contrast the logos of the French Open and the US Open tennis championships. & Text-Text, Text-Table, Text-Image, Table-Table, Table-Image, Image-Image & 227 \\ \hline Numerical & Requires numerical calculations applied to facts extracted from the sources. & How many years after the 1st Academy Awards did Dustin Hoffman receive his first nomination? & Text, Table, Text-Text, Text-Table, Table-Table & 208 \\ \hline Compound & Two loosely connected information extraction questions, separated by ``and''. & What is the scale used for passer rating in the NFL, and who is the all-time passing yards leader in professional football league history? & Text-Text, Text-Table, Text-Image, Table-Table, Table-Image, Image-Image & 206 \\ \hline Multi-hop & Requires first resolving an implicit sub-question, the answer to which is used to resolve the full question. & What color is the background of the flag of the location where the Echigo-Tsumari Art Triennial is held? & Text-Text, Text-Table, Text-Image, Table-Table, Table-Image & 225 \\ \hline
    \end{tabular}
    \caption{\textbf{Summary of our SMMQG-generated Wikipedia QA dataset.} This dataset consists of 1024 QA pairs, spanning five question styles and all pairwise modality combinations. Examples are taken directly from our dataset.}
    \label{tab:question_styles}
\end{table*}

\section{Experiments}\label{sec:experiment_i}

\subsection{Building a Synthetic Multimodal Wikipedia QA Dataset}\label{sec:dataset}

We use SMMQG to build a QA dataset over Wikipedia documents. We use this dataset to facilitate experiments (1) demonstrating the utility of SMMQG as a framework for generating style- and modality-specific datasets (Section \ref{sec:experiment_i}) and (2) verifying that SMMQG produces high quality data (Section \ref{sec:experiment_ii}). We use the text passages, tables and images gathered by \citet{talmor2021multimodalqa} for MMQA as our sources. These sources include approximately 57,000 captioned images, 232,000 text passages and 12,000 tables, and were scraped from the 2020-01-01 English Wikipedia dump. 

We preprocess the dataset by removing text passages with lengths of less than 200 characters, as we find that short passages rarely contain enough information to generate good questions. We generate a total of 1024 QA samples across five different question styles covering all pairwise modality combinations.\footnote{SMMQG can be used to generate multimodal questions with three or more question sources \cite{luo-etal-2023-unifying}. We leave exploration of this to future work.} These are summarised in Table \ref{tab:question_styles}, and further examples are detailed in Appendix \ref{sec:appendix_dataset_examples}. We choose these five question styles because (1) they test for a diverse set of reasoning abilities (2) they are well-represented across the QA literature \cite{khashabi-etal-2018-looking, rajpurkar2016squad, yang2018hotpotqa, dua2019drop}.

We also generate an additional dataset from different source documents to demonstrate that SMMQG is compatible with more domain-specific sources. See Appendix \ref{sec:appendix_bio_dataset} for details.

\subsection{Retriever and QA Model Evaluation}\label{sec:eval_experiments}
\begin{table*}[h]
    \small
    \centering
    \begin{tabular}{p{0.09\linewidth} p{0.055\linewidth} | c c c c c}
    \hline
     & & Info Extraction & Compare Contrast & Numerical & Compound & Multi-hop \\ \hline
     BM25 & top-5 & 53.2 & 37.7 & 56.8 & 37.4 & 35.8 \\
        & top-10 & 56.6 & 44.5 & 63.6 & 42.0 & 39.3 \\
     E5 & top-5 & 65.8 & 65.9 & 76.9 & 52.4 & 44.4 \\
        & top-10 & \textbf{67.8} & \textbf{72.2} & \textbf{82.6} & \textbf{57.5} & \textbf{50.9} \\
     OpenCLIP & top-5 & 61.1 & 38.3 & 51.6 & 36.9 & 18.0 \\
        & top-10 & 67.1 & 48.9 & 59.5& 43.3 & 24.0 \\ \hline
     \noalign{\vskip 4mm}    
     \hline
     \multicolumn{2}{l|}{Vicuna-7b + LLaVA-7b} & 81.6 & 35.9 & 13.3 & 52.9 & 49.3 \\
     \multicolumn{2}{l|}{Vicuna-13b + LLaVA-13b} & \textbf{89.1} & \textbf{55.5} & 33.2 & \textbf{73.5} & 60.2 \\
     \multicolumn{2}{l|}{Qwen-Chat + Qwen-VL-Chat} & 87.5 & 48.0 & \textbf{42.7} & 67.7 & \textbf{65.6} \\ \hline
     \multicolumn{2}{l|}{Gemini Pro 1.0} & \textbf{96.3} & 51.4 & 64.0 & 89.8 & 74.3 \\
     \multicolumn{2}{l|}{Claude 3 Haiku} & 90.4 & 54.6 & 44.8 & 75.7 & 54.9 \\
     \multicolumn{2}{l|}{Claude 3 Sonnet} & 93.0 & 67.0 & 63.3 & 79.2 & 72.7 \\
     \multicolumn{2}{l|}{Claude 3 Opus} & 96.1 & \textbf{81.3} & \textbf{77.7} & \textbf{90.8} & \textbf{88.7} \\ \hline
     \multicolumn{2}{l|}{GPT-4-Turbo} & 99.3 & 89.9 & 85.3 & 96.8 & 96.2 \\ \hline
    \end{tabular}
    \caption{\textbf{Retrieval and QA evaluation results by question style}. The \textit{top} subtable contains retrieval recall@5 and recall@10. The \textit{bottom} subtable contains GPT-4-Turbo-judge scores for QA, where the top section contains results for open-source models, the middle contains results for proprietary models, and the bottom contains results for GPT-4-Turbo. We denote in \textbf{bold} the best models in each category. For retrieval, E5 achieves the best performance across all styles. Amongst open-source QA models, no single model dominates, while for proprietary QA models excluding GPT-4-Turbo, Claude 3 Opus demonstrates the best performance.}
    \label{tab:eval_results_question_styles}
\end{table*}

\begin{table*}[h]
    \small
    \centering
    \begin{tabular}{p{0.09\linewidth} | p{0.055\linewidth} | c c c c c c c c c}
    \hline
     & & Txt & Tab & Im & Txt-Txt & Txt-Tab & Txt-Im & Tab-Tab & Tab-Im & Im-Im \\ \hline
     BM25 & top-5 & 92.9 & 50.0 & 12.0 & 76.9 & 39.8 & 43.4 & 20.0 & 20.2 & 15.4 \\
          & top-10 & 95.2 & 54.5 & 14.0 & 86.1 & 45.5 & 46.5 & 26.5 & 24.5 & 18.4 \\
     E5 & top-5 & 98.8 & 79.5 & 20.0 & 91.5 & 60.2 & 52.3 & 51.0 & 34.6 & 24.3 \\
        & top-10 & \textbf{100.0} & \textbf{81.8} & 22.2 & \textbf{96.3} & \textbf{70.3} & \textbf{56.2} & \textbf{60.2} & \textbf{38.5} & 26.5 \\
     OpenCLIP & top-5 & 78.6 & 30.7 & 84.0 & 55.8 & 28.0 & 36.4 & 21.1 & 19.2 & 39.7 \\
        & top-10 &  85.7 & 36.4 & \textbf{90.2} & 63.6 & 36.2 & 46.1 & 28.6 & 23.6 & \textbf{49.3} \\ \hline
     \noalign{\vskip 4mm}    
     \hline
     \multicolumn{2}{l|}{Vicuna-7b + LLaVA-7b} & 65.9 & 43.5 & \textbf{74.0} & 52.6 & 28.0 & 60.1 & 12.5 & 43.2 & 46.8 \\
     \multicolumn{2}{l|}{Vicuna-13b + LLaVA-13b} & 77.6 & 60.2 & \textbf{74.0} & \textbf{70.9} & \textbf{63.8} & 61.9 & 50.4 & 43.7 & \textbf{47.1} \\
     \multicolumn{2}{l|}{Qwen-Chat + Qwen-VL-Chat} & \textbf{82.3} & \textbf{66.7}  & 65.0 & 67.0 & 56.2 & \textbf{69.6} & \textbf{51.3} & \textbf{50.5} & 30.9 \\ \hline
     \multicolumn{2}{l|}{Gemini Pro 1.0} & 84.4 & 71.3 & \textbf{97.0} & 78.6 & 57.3 & 82.9 & 59.1 & \textbf{84.6} & 80.4 \\
     \multicolumn{2}{l|}{Claude 3 Haiku} & 78.4 & 61.6 & 87.8 & 61.3 & 42.4 & 75.7 & 36.2 & 63.1 & 80.1 \\
     \multicolumn{2}{l|}{Claude 3 Sonnet} & 92.7 & 75.7 & 86.7 & 86.5 & 65.8 & 76.3 & 51.8 & 72.1 & 80.4 \\
     \multicolumn{2}{l|}{Claude 3 Opus} & \textbf{96.1} & \textbf{91.2} & 92.0 & \textbf{90.9} & \textbf{81.9} & \textbf{84.9} & \textbf{75.9} & 83.7 & \textbf{86.0} \\ \hline
     \multicolumn{2}{l|}{GPT-4-Turbo} & 97.4 & 92.7 & 98.0 & 96.5 & 94.1 & 93.4 & 86.2 & 93.3 & 89.0 \\ \hline
    \end{tabular}
    \caption{\textbf{Retrieval and QA evaluation results by modality}. The \textit{top} subtable contains retrieval recall@5 and recall@10. The \textit{bottom} subtable contains GPT-4-Turbo-judge scores for QA, where the top section contains results for open-source models, the middle contains results for proprietary models, and the bottom contains results for GPT-4-Turbo. We denote in \textbf{bold} the best models in each category. For retrieval, OpenCLIP performs best for pure image retrieval, while E5 performs best for other modalities. Amongst open-source QA models, no single model dominates, while for proprietary QA models excluding GPT-4-Turbo, Claude 3 Opus generally demonstrates the best performance.}
    \label{tab:eval_results_modalities}
\end{table*}

We use our dataset to evaluate the performance of three retrievers\footnote{We use ChromaDB \cite{trychromaHomeChroma} as the vector database for storing and retrieving embeddings.} and eight LLM + LMM combinations. For retrieval, we evaluate BM25, E5-Large \cite{wang2024text} (both text-based) and OpenCLIP \cite{cherti2022reproducible} -- an open-source implementation of CLIP offering improved performance -- using recall@5 and recall@10 as our metric. For the text-based retrievers, we rely on captions and verbalisations for retrieving over images. 

We evaluate the LLM + LMM combinations by conditioning answer generation on the SMMQG questions and question sources\footnote{This evaluation setup is easier than end-to-end evaluation where the retrieved sources are used for QA, but has the advantage of allowing for independent evaluation of the retrieval and QA components.} and then scoring the predictions against the answers using GPT-4-Turbo as a judge, which has been shown to produce judgements that correlate strongly with human judgements \cite{zheng2023judging, kim2024prometheus}. The judge is asked to score the predicted answer on a three-point scale given the SMMQG-generated answer and the question sources. For QA, we use the LMM when the question sources contain at least one image; otherwise we use the LLM. An example of the judge prompt is provided in Appendix \ref{sec:appendix_judge}. In addition to GPT-4-Turbo-judge scores, we also report GPT-3.5-Turbo-judge \citep{brown2020language}, BERTScore \citep{zhang2020bertscoreevaluatingtextgeneration} and ROUGE \citep{lin-2004-rouge} scores. We include these results in Appendix \ref{sec:appendix_additional_judge}.

We assess the open-source model combinations Vicuna-7b-v1.5 + LLaVA-v1.5-7b, its 13b counterpart \cite{zheng2023judging, liu2023visual}, and Qwen-Chat + Qwen-Chat-VL \cite{bai2023qwen, bai2023qwenvl}. The LLaVA-v1.5 models are LMMs finetuned from Vicuna-v1.5 \cite{peng2023instruction}, and both Qwen models are 7B parameter models finetuned from Qwen-LM \cite{bai2023qwen}. We also assess the proprietary multimodal models GPT-4-Turbo \cite{openai2024gpt4}, Gemini Pro 1.0 \cite{geminiteam2024gemini} and the Claude 3 family \cite{claude3}, which are trained to process text-only inputs in addition to multimodal inputs and so can act as both the LLM and LMM. See Appendix \ref{sec:appendix_model_details} for further details.

\textbf{Evaluation Results}\quad Retriever evaluation results are shown in the top subtables of Tables \ref{tab:eval_results_question_styles} and \ref{tab:eval_results_modalities}. We find that the text-based E5 retriever outperforms BM25 and OpenCLIP across all question styles, and is especially strong on abstractive styles (numerical, compare contrast, multi-hop), although we stress that our evaluation process is likely biased towards E5 as it was used in the question generation process itself. We also find that OpenCLIP significantly outperforms text-based retrieval for pure image retrieval questions, but underperforms text-based retrieval for all other modalities. Our findings underscore the importance of using robust, general-purpose multimodal embedders that are capable of performing consistently across diverse modalities, and we hope that our work can be used to support the development of such models.

The QA evaluation results are shown in the bottom subtables of Tables \ref{tab:eval_results_question_styles} and \ref{tab:eval_results_modalities}. To start, we see that GPT-4-Turbo outperforms all other models across all question styles and modalities. This strength may be attributable to (1) GPT-4-Turbo being genuinely strong (2) GPT-4-Turbo being the evaluation data generation model, which filters out many questions that the model is unable to answer (3) GPT-4-Turbo being itself used as the judge, as LLM judges favour their own outputs \cite{Koo2023BenchmarkingCB}. 

Amongst the other models, Claude 3 Opus generally performs best, although further analysis yields novel style- and modality-specific insights. Firstly, Gemini Pro 1.0 performs similarly to Claude 3 Opus on extractive question styles (information extraction, compound), but is weaker on abstractive styles, where it is comparable to the smaller Claude 3 models. Secondly, the gap between open-source and proprietary models is small for extractive question styles but large for abstractive styles, with open-source models being especially poor on numerical and compare contrast questions. Comparing open-source models, we find that Qwen-Chat + Qwen-VL-Chat performs better on numerical and multi-hop questions than Vicuna-13b + LLaVA-13b despite being significantly smaller. 

On the modality front, we learn that Gemini Pro 1.0 performs strongly for questions containing image sources, and is stronger than even Claude 3 Opus for unimodal image questions, although it suffers from weaker text and table reasoning abilities. Claude 3 Haiku, meanwhile, is surprisingly poor at table QA (comparable to open-source models), but makes up for this with superior image reasoning capabilities.

In summary, we demonstrate that SMMQG can generate question style- and modality-specific evaluation datasets. Such datasets reveal important insights into the style- and modality-specific strengths and weaknesses of retrievers and QA models that would otherwise remain hidden. 

\section{Assessing Data Quality}\label{sec:experiment_ii}

\subsection{MMQA}\label{sec:mmqa}
In the following sections, we assess the quality of our SMMQG-generated dataset and use MMQA \cite{talmor2021multimodalqa} as a reference to better understand the significance of our results. MMQA is a crowdsourced benchmark constructed over the same documents as ours. It contains both uni- and cross-modal questions, and provides long-form answers. 

We use MMQA as our reference dataset for three reasons. Firstly, the modalities present in MMQA overlap exactly with our modalities of interest, unlike datasets such as WebQA \cite{chang2022webqa} and OK-VQA \cite{marino2019okvqa}, which contain only text and images. Secondly, MMQA is built on source documents spanning a diverse set of domains (Wikipedia), unlike domain-specific datasets such as BioASQ \cite{krithara2023bioasq} (biomedical) and MMMU \cite{yue2023mmmu} (exams and textbooks). Using it as a reference is therefore likely to yield more generalisable results. Lastly, the \textit{single modality} and \textit{compose}\footnote{\textit{Single modality} questions are the TextQ, TableQ and ImageQ and \textit{compose} questions the Compose(TextQ, TableQ), Compose(TableQ, TextQ), Compose(ImageQ, TextQ) and Compose(ImageQ, TableQ) question types from MMQA.} questions from MMQA are stylistically very similar to the info extraction and multi-hop questions from our SMMQG dataset; the presence of overlapping question styles enables direct comparison between datasets. 

\subsection{Human Study}\label{sec:human_study}

\begin{table*}[ht]
    \small
    \centering
    \begin{tabular}{p{0.15\linewidth} | c c c c c}
    \hline
      & Q. Fluency & Q. Style Faithfulness & Source Relevance & Answerability & A. Correctness \\ \hline
     SMMQG & 4.53 & 98.3 & 93.0 & 94.7 & 92.7 \\
     MMQA & 3.68 & 96.7 & 85.8 & 85.8 & 80.0 \\ \hline
     \centering $\Delta$ & \textbf{+0.85} & +1.6 & +7.2 & \textbf{+8.9} & \textbf{+12.7} \\ 
     \centering $p$-value & <0.001* & 0.77 & 0.07 & 0.02* & 0.001* \\ \hline
    \end{tabular}
    \caption{\textbf{Human study results}. We denote statistically significant $\Delta$ in \textbf{bold} and $p$-values with $p \leq 0.05$ using *. These results show that our SMMQG dataset quality is on par with (and sometimes exceeds) the quality of MMQA.}
    \label{tab:human_study_main}
\end{table*}

We conduct a human study to directly evaluate the quality of SMMQG-produced data. We randomly draw 300 samples from our dataset, along with 60 single modality and 60 compose samples from the train set of MMQA, ensuring even distribution over modalities. We combine the samples and ask crowdworkers to rate them along five metrics:
\begin{itemize}[itemsep=0pt, leftmargin=5mm]
    \item \textbf{Question Fluency} (5-likert): How fluent is the question?
    \item \textbf{Question Style Faithfulness} (Yes/No): Is the question faithful to the question style? 
    \item \textbf{Source Relevance} (Yes/No): Are all the sources relevant to answering the question?
    \item \textbf{Answerability} (Yes/No): Is the question answerable using only information in the sources? 
    \item \textbf{Answer Correctness} (Yes/No): Is the answer correct given the sources? 
\end{itemize}

We compute the average rating for each metric for each dataset and report the results in Table \ref{tab:human_study_main}. See Appendix \ref{sec:appendix_human_study} for further details on our human study methodology and for style-specific results. We employ the Mann-Whitney U test\footnote{This non-parametric test compares differences between two independent groups when the dependent variable is either ordinal or continuous but not normally distributed.} (for Question Fluency) and Fisher's exact test\footnote{This test is used to determine if there are nonrandom associations between two categorical variables. In our case, the variables are dataset (SMMQG/MMQA) and human judgement for a given metric (Yes/No).} (for the remaining metrics) to determine statistical significance. Our findings are as follows:

\textbf{SMMQG achieves high question style faithfulness.} This finding suggests that SMMQG can reliably produce questions with styles based on user specifications. 

\textbf{SMMQG questions are more fluent than MMQA questions across comparable styles.} Any results we obtain better reflect the true reasoning capabilities of the model, as there is less interference caused by poor phrasing. However, we do not assess the ability of the model to handle poorly-phrased questions, which may reflect real user queries \cite{kwiatkowski-etal-2019-natural}. 

\textbf{SMMQG question sources are highly relevant.} One source of error in SMMQG arises when the question-generation model selects question sources that the generated question is not based on. Our human study results address this concern.

\textbf{SMMQG questions are highly likely answerable given the question sources.} Another potential source of error arises when the question-generation model generates questions that are not answerable given its own selected question sources. Our study shows that SMMQG questions are actually statistically significantly more likely answerable than MMQA questions. 

\textbf{SMMQG answers are highly likely to be correct.} A high level of answer correctness reduces the noise associated with incorrect labels leading to incorrect assessments of the predicted answers, and our SMMQG answers are statistically significantly more likely to be correct.

\subsection{Measuring Concurrence}\label{sec:concurrence}

\begin{table}[h]
    \small
    \centering
    \begin{tabular}{p{0.15\linewidth} | c c}
    \hline
      & $\tau$ & $p$-value \\ \hline
     Retrieval & 0.87 & 0.02* \\
     QA  & 0.86 & 0.002* \\ \hline
    \end{tabular}
    \caption{\textbf{Concurrence of SMMQG for retrieval and QA.} We report Kendall's tau and its associated $p$-values on the ranked list of retrieval and QA model evaluation results and find strong concurrence ($\tau > 0.8$). We use * to denote statistical significance ($p \leq 0.05$).}
    \label{tab:concur}
\end{table}

We compute the \textit{concurrence} \cite{liu2023question} between our SMMQG dataset and MMQA. The motivation for this is as follows: if we assume that MMQA is a useful evaluation dataset, and if our SMMQG dataset discriminates between models in the same way as MMQA, then we can reasonably conclude that our SMMQG dataset is also a useful evaluation dataset \cite{viswanathan2023prompt2model}. 

We randomly draw 150 information extraction and 150 multi-hop samples from our SMMQG dataset, evenly distributed over modalities. We also randomly draw 150 single modality and 150 compose samples from MMQA, again distributed evenly over modalities. We then run evaluation using the methodology described in Section \ref{sec:eval_experiments} (see Appendix \ref{sec:appendix_concur_study} for these results) and calculate concurrence by computing Kendall's tau on the ranked lists of these results. We report our findings in Table \ref{tab:concur}. 

We find that MMQA and SMMQG strongly concur ($\tau > 0.8$, \citet{liu2023question}) for both retrieval and QA. This implies that our SMMQG dataset can be used in-place of MMQA to discriminate between models (at least for the two common question styles), further validating its quality.

\section{Related Work}
\textbf{Multimodal QA Benchmarks}\quad Many existing benchmarks rely on human annotators to handcraft questions and answers over fixed sets of documents. MMQA \cite{talmor2021multimodalqa} curate question-answer pairs by combining compositional question templates with crowdsourcing. This limits the complexity and diversity of generated questions \cite{chen2022murag}. MMMU \cite{yue2023mmmu}, which was constructed from college-level education material, was costly to curate, requiring the input of over 50 individuals. Other human-crafted multimodal QA benchmarks include WebQA \cite{chang2022webqa}, BioASQ \cite{krithara2023bioasq}, ScienceQA \cite{lu2022learn}, InfoSeek \cite{chen2023pretrained} and OK-VQA \cite{marino2019okvqa} 

\textbf{Synthetic QA Generation}\quad There exists a large body of work leveraging synthetic data to train and evaluate text-only QA systems. \citet{puri2020training} generate extractive QA data and use this to train a BERT-based \cite{devlin2019bert} model for QA. \citet{shakeri2020endtoend} generate QA pairs using BART \cite{lewis2019bart} and use this for domain adaptation. \citet{pan2020unsupervised} propose a multi-step process to generate multi-hop questions, while \citet{es2023ragas} create a synthetic evaluation dataset for text-based RAG evaluation. Synthetic question generation has also been used to evaluate the quality of text summaries \cite{Durmus_2020, wang2020asking}. Note that, unlike SMMQG, none of the works discussed above allow for fine-grained control of question styles.

\textbf{Synthetic Multimodal QA Generation}\quad Synthetic data generation has been studied in the context of visual QA (VQA). MultiQG-TI \cite{wang2023multiqg} utilize an image-to-text model and OCR to create text descriptions of images that are combined with text passages and then passed to an LLM for QA generation. \citet{patel2020generating} build a diverse synthetic QA dataset from images and their associated metadata using an image-to-text model. These datasets concern QA over images and possibly image-text pairs only, and do not address text-only or table modalities. Furthermore, they do not enable control over question styles.

\section{Conclusion}

We introduce SMMQG, a framework for generating synthetic multimodal questions and answers grounded in multimodal documents that adhere to user-specified question styles and modalities. We demonstrate that an SMMQG-generated evaluation dataset is able to reveal novel style- and modality-specific insights into the performance of state-of-the-art retrievers, LLMs and LMMs. Through a human study and by measuring dataset concurrence, we also show that the quality of data generated by SMMQG is on-par with the quality of data from crowdsourced benchmark MMQA and that our dataset can be used in place of MMQA for model selection. We hope that SMMQG will enable automatic, large-scale and tailored evaluation of MMRAG systems, thereby facilitating their adoption in practical applications.

\section*{Limitations}

In our work, we evaluate SMMQG on Wikipedia documents and use GPT-4-Turbo to generate questions belonging to five question styles. SMMQG performance may differ when these variables are altered. The impact of new question styles and documents on SMMQG performance depends largely on how well GPT-4-Turbo is able to reason over these new question styles and understand these documents. This highlights a limitation of our work: we presuppose the existence of a model capable of reasoning over our question style and understanding our document of choice. Nonetheless, when such a model does exist, SMMQG enables us to generate synthetic data that can be used to evaluate other, possibly weaker models. Another related limitation of our work is that we only assess the viability of E5-Large and GPT-4-Turbo as the SMMQG retriever and question-generation model, and it is possible that new challenges arise when other models are used. Another limitation is that we limit our study to use of SMMQG data for evaluation, even though it is possible to use it for training. We encourage further research in this direction, but we cannot yet claim that SMMQG data is appropriate for model training. Finally, our work relies on the existence of a high-quality set of unpaired data sources (in our experiments, this consists of images, text passages, and tables). This may not be an appropriate assumption in all situations; we did not test the ability of our dataset generation method to generalize to noisy data sources.

\section*{Potential Risks}

LLMs and LMMs are known to generate false, harmful and biased material. As SMMQG leverages LLMs and LMMs, it may potentially generate false, harmful and biased questions and answers. We also note that we have only explored the use of SMMQG for assessing the QA performance of MMRAG, and not its alignment to human values and preferences. MMRAG systems may excel on SMMQG-generated datasets but nonetheless be misaligned.

\bibliography{custom}


\appendix

\onecolumn\section{SMMQG Dataset Examples}
\label{sec:appendix_dataset_examples}

In this section, we provide examples of SMMQG-generated questions. Examples are cherry-picked to showcase questions of various question styles and modalities.

\subsection{Info Extraction, Image}
\fbox{
    \parbox{0.98\textwidth}{
        \textbf{Question:} What musical instrument is Anita Cochran holding on the cover of her album "Back to You"? \\ \\
        \textbf{Answer:} Anita Cochran is holding a guitar on the cover of her album "Back to You". \\ \\
        \textbf{Question Source 1:} \\ \\
        \begin{minipage}{\linewidth}
            \centering
            \includegraphics[width=0.3\textwidth]{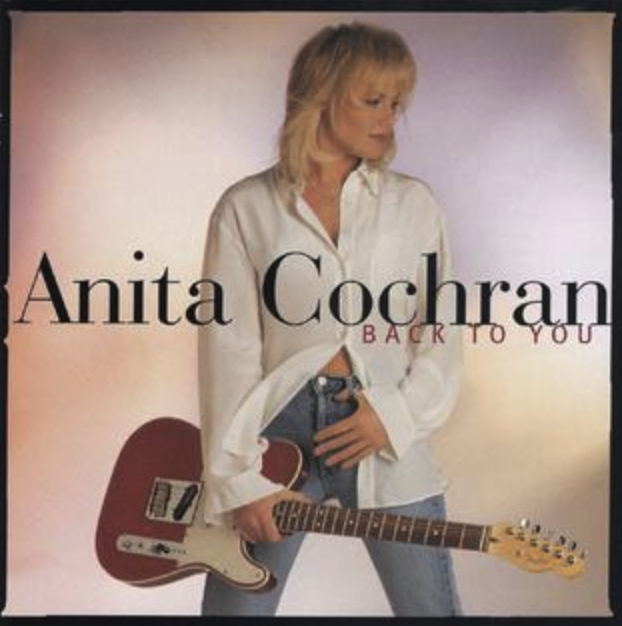}
        \end{minipage} \\ \\
        \textbf{Image Caption:} Back to You (album) \\
        \textbf{Image Verbalisation:} Anita Cochran, a country music singer, poses with her guitar for her album cover. The image features her holding the guitar and wearing a white shirt, showcasing her talent and passion for music.
    }
}

\subsection{Numerical, Table}
\fbox{
    \parbox{0.98\textwidth}{
        \textbf{Question:} What is the average position Pavlo Cherepin finished in during the 2014 rally season? \\ \\
        \textbf{Answer:} To calculate the average position, we add the positions from the 2014 season: 1 (Monte Carlo Rally) + 1 (Rally Mexico) + 3 (Rally d'Italia) + 1 (Rally Australia) and then divide by the number of events, which is 4. The calculation is (1+1+3+1)/4 = 6/4 = 1.5. Therefore, the average position Pavlo Cherepin finished in during the 2014 rally season is 1.5. \\ \\
        \textbf{Question Source 1:} \\ \\
        \begin{minipage}{\linewidth}
            \centering
            \includegraphics[width=0.7\textwidth]{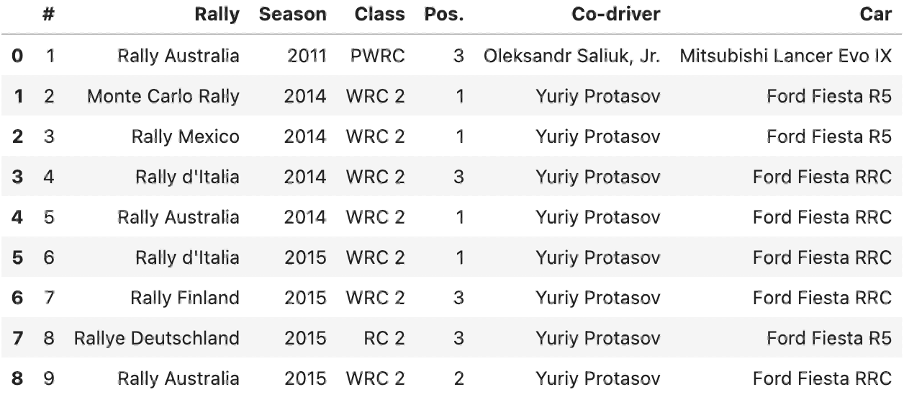}
        \end{minipage} \\ \\
        \textbf{Table Title:} Pavlo Cherepin
    }
}

\subsection{Multi-hop, Text-Image}
\fbox{
    \parbox{0.98\textwidth}{
        \textbf{Question:} What colors are used in the logo of the NBA team that is based in Washington, D.C. and competes in the Eastern Conference Southeast Division? \\ \\
        \textbf{Answer:} The NBA team that is based in Washington, D.C. and competes in the Eastern Conference Southeast Division is the Washington Wizards. The Washington Wizards logo uses red, white, blue, and silver colors. Therefore, the colors used in the logo of the NBA team based in Washington, D.C. and competing in the Eastern Conference Southeast Division are red, white, blue, and silver. \\ \\
        \textbf{Question Source 1:} Washington Wizards: The Washington Wizards are an American professional basketball team based in Washington, D.C. The Wizards compete in the National Basketball Association (NBA), as a member of the league's Eastern Conference Southeast Division. The team plays its home games at the Capital One Arena, in the Chinatown neighborhood of Washington, D.C. \\ \\
        \textbf{Question Source 2:} \\ \\
        \begin{minipage}{\linewidth}
            \centering
            \includegraphics[width=0.2\textwidth]{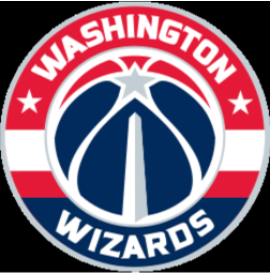}
        \end{minipage} \\ \\
        \textbf{Image Caption:} Washington Wizards \\
        \textbf{Image Verbalisation:} Washington Wizards Basketball Team Logo
    }
}

\subsection{Compare Contrast, Text-Text}
\fbox{
    \parbox{0.98\textwidth}{
        \textbf{Question:} Compare and contrast the sculptures "The Spirit of Polonia" and "On Watch" in Milwaukee.  \\ \\
        \textbf{Answer:} Both "The Spirit of Polonia" and "On Watch" are public sculptures located in Milwaukee, but they differ in their themes and representations. "The Spirit of Polonia" is a stainless steel sculpture with three rings symbolizing harmony, unity, and infinity, and a brass sphere representing earth, emphasizing abstract concepts. In contrast, "On Watch" is a life-size bronze sculpture depicting a policeman, a firefighter, and a child, focusing on the human figures and their roles as guardians of public safety. Additionally, "The Spirit of Polonia" is situated in front of the Milwaukee County Courthouse and includes a concrete pool, while "On Watch" is administered by the City of Milwaukee and features inscriptions and a plaque with dedications. \\ \\
        \textbf{Question Source 1:} On Watch: "On Watch" is a life-size bronze sculpture consisting of three figures. A Caucasian policeman and an African-American firefighter stand back-to-back holding a child wrapped in a blanket. The inscription on the southwest corner of the sculpture reads: WANNER 90. The inscription on the northwest corner reads: WACO FDR. There is a bronze plaque on the concrete base which reads: On Watch/ Dedicated to the guardians of public safety in the city of Milwaukee John Norquist, mayor Milwaukee Art Commission, David M. Wanner Sculptor 1990. The work is administered by the City of Milwaukee. \\ \\
        \textbf{Question Source 2:} "The Spirit of Polonia": This nine foot, five inch stainless steel sculpture has three rings meaning harmony, unity and infinity. While the brass sphere represents earth. Each ring is a different size having one inside the other, therefore having each one get smaller, then the "globe" is the smallest. These sculpture is surrounded by a sixteen-foot, five inch concrete pool. Both are in front of the Milwaukee County Courthouse.
    }
}

\subsection{Compound, Text-Table}
\fbox{
    \parbox{0.98\textwidth}{
        \textbf{Question:} What is the home stadium of the Miami Dolphins, and what was the attendance when the Baltimore Colts played against the Miami Dolphins in 1975? \\ \\
        \textbf{Answer:} The home stadium of the Miami Dolphins is Hard Rock Stadium, and the attendance when the Baltimore Colts played against the Miami Dolphins in 1975 was 61,986 on Week 10 and 59,398 on Week 13. \\ \\
        \textbf{Question Source 1:} Hard Rock Stadium: Hard Rock Stadium is a multipurpose football stadium located in Miami Gardens, Florida, a city north of Miami. It is the home stadium of the Miami Dolphins of the National Football League (NFL). Hard Rock Stadium also plays host to the Miami Hurricanes football team during their regular season. The facility also hosts the Orange Bowl, an annual college football bowl game. It was the home to the Florida Marlins of Major League Baseball (MLB) from 1993 to 2011. \\ \\
        \textbf{Question Source 2:} \\ \\
        \begin{minipage}{\linewidth}
            \centering
            \includegraphics[width=0.8\textwidth]{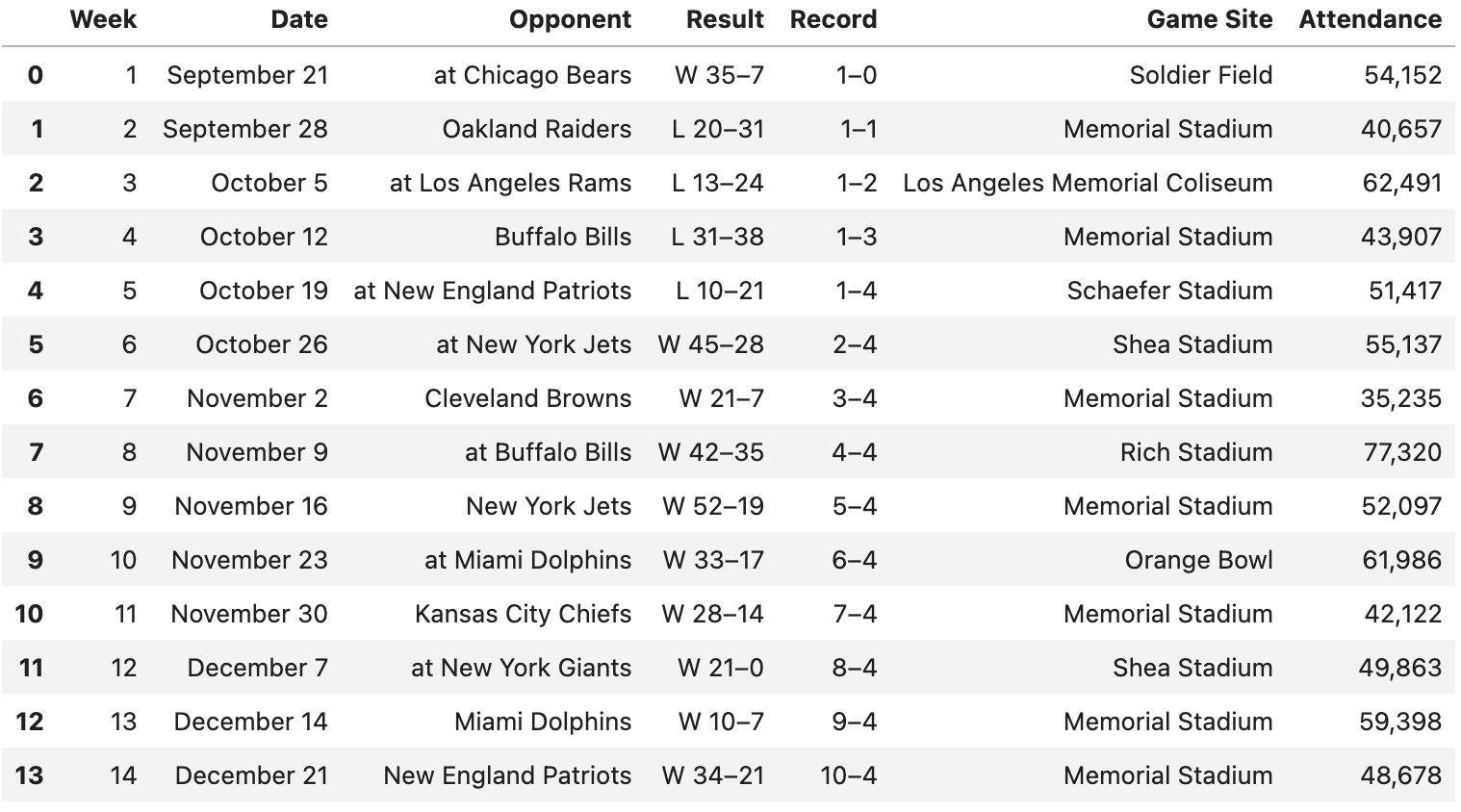}
        \end{minipage} \\ \\
        \textbf{Table Title:} 1975 Baltimore Colts season
    }
}

\onecolumn\section{Image Verbalisation}
\label{sec:appendix_verbalisation}

All image verbalisations used for our experiments were created with the \texttt{llama-2-13b-chat-lightning-preview} version of LLaVA, which was the most up-to-date LLaVA model at the time the verbalisations were created.

\subsection{Image Verbalisation Prompt} 
We pass the image along and any associated captions to LLaVA and use the following prompt to generate the image verbalisation: \\ \\
\fbox{
    \parbox{0.98\textwidth}{\texttt{Here is the image caption: \{caption\}. \\Here is an image. Describe the contents of the image, taking into account all portions. Try and be descriptive.}
    }
} \\

\subsection{Image Verbalisation Examples} 

\begin{minipage}[t]{0.35\textwidth}
\strut\vspace*{-\baselineskip}\newline
    \centering
    \includegraphics[width=\textwidth]{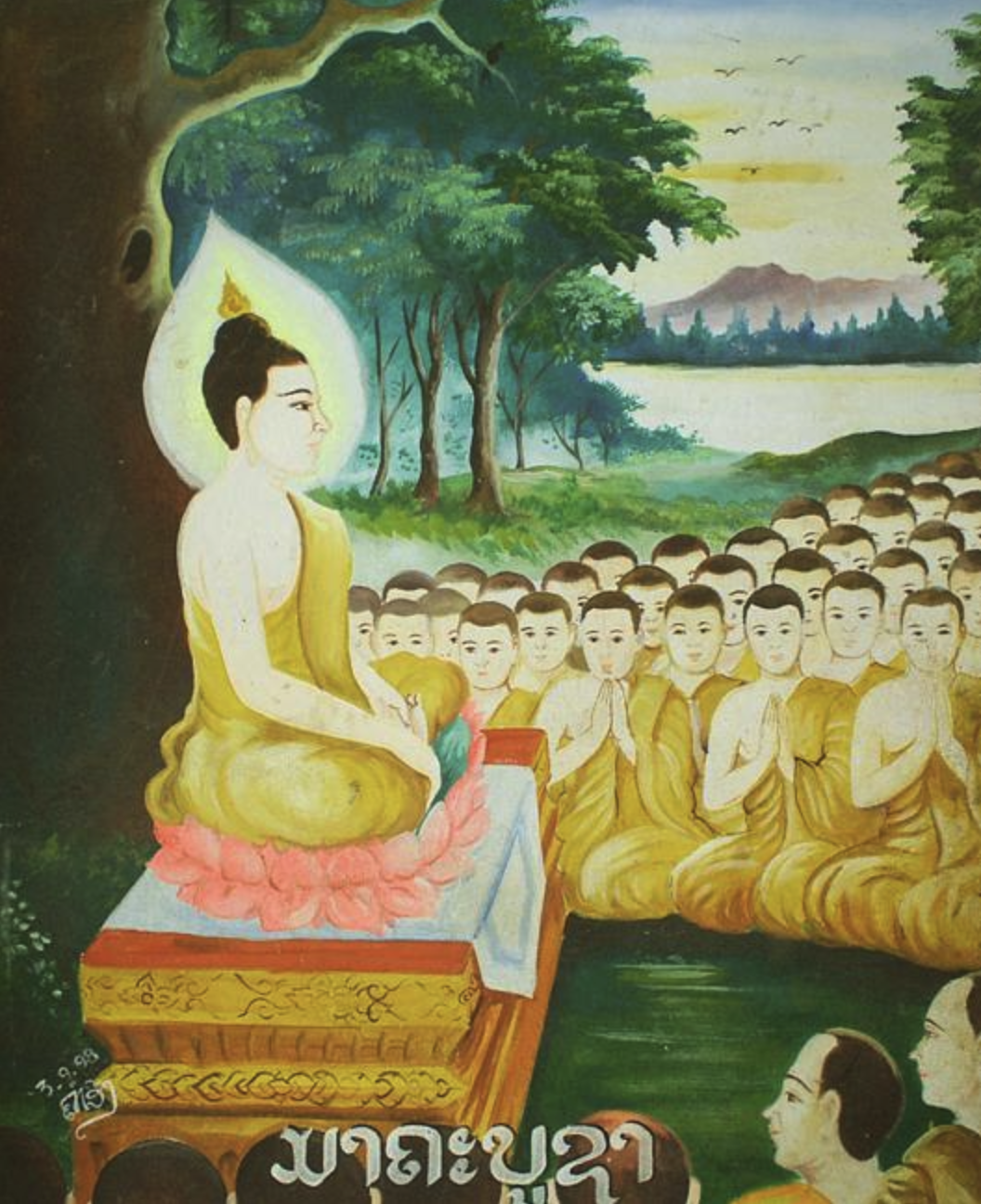} 
\end{minipage}%
\hspace{0.05\linewidth}
\begin{minipage}[t]{0.5\textwidth}
\strut\vspace*{-\baselineskip}\newline
    \textbf{Image Caption:} Magha Puja \\ \\
    \textbf{Image Verbalisation:} In this painting, a large group of monks is gathered around a seated Buddha, listening intently to his teachings. The scene takes place in a serene outdoor setting, with a tree in the background. The image captures the essence of spirituality and the importance of learning from one's spiritual leader.
\end{minipage} \vspace{2.0em} \\

\begin{minipage}[t]{0.35\textwidth}
\strut\vspace*{-\baselineskip}\newline
    \centering
    \includegraphics[width=\textwidth]{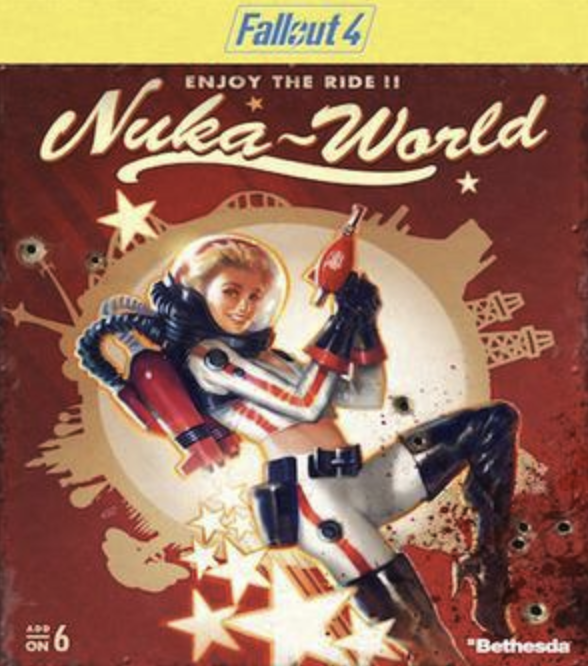} 
\end{minipage}%
\hspace{0.05\linewidth}
\begin{minipage}[t]{0.5\textwidth}
\strut\vspace*{-\baselineskip}\newline
    \textbf{Image Caption:} Fallout 4: Nuka-World \\ \\
    \textbf{Image Verbalisation:} A woman in a space suit is enjoying a thrilling ride on a roller coaster, with a bottle of soda in her hand. This image captures the excitement and enjoyment of the theme park experience.
\end{minipage}

\onecolumn\section{SMMQG Prompts}
\label{sec:appendix_smmqg_prompts}

\subsection{Entity Extraction Prompt}

Here we provide the entity extraction prompts from Step~2 of SMMQG. We choose \{num\_entities\} to be 1 in our experiments, although this can be increased to further increase question diversity. \\ \\
\fbox{
    \parbox{0.98\textwidth}{
        \texttt{Identify up to \{num\_entities\} key themes or entities present in the passage. Make sure any entities you return are general and widely-known.\\ \\ <few shot examples> \\ \\ Passage: \{passage\} \\ \\Entities and terms:}
    }
} \\

\subsection{Question Generation Prompts}
Here we provide the question generation prompts for Step~4 of SMMQG. For inputs containing text or tables only, the prompt is:\\ \\
\fbox{
    \parbox{0.98\textwidth}{
        \texttt{You are given a question style description, which describes the characteristics of a specific style of reading comprehension question, and some examples of questions of this style. You are also provided with modality requirements. Your task is to generate a question following the style specified in the question style description based on the input passages. Also generate an answer to the question and citations of the passages the answer is based on.\\ \\RULES\\1. The question you generate MUST be based on one or more of the provided source passages.\\2. The modality requirements constrain the passages you can choose as the source passages. For example, if the modality requirement is 1 text, 1 table, your question MUST be based on 1 text and 1 table passage exactly.\\3. The question should not be answerable if any chosen source passage is removed.\\4. If no modality requirements are given, you may generate questions based on any number of passages of any modality.\\5. Do not mention the passage number in the question. Also do not explicitly mention the table, image or text.\\6. Generate a natural sounding question that a reasonable human being might ask.\\7. Produce a response in the following format: <question> | <answer> | <citation>. The citation should be a reference to all the source passages chosen.\\8. Closely follow the template question description. If you cannot do this, say None.\\9. If you cannot abide by any of the rules above, say None. It is preferrable for you to say None than to risk breaking the rules.\\ \\ \{style\_prompt\} \\ \\ <few shot examples> \\ \\ Passages: \{enumerated\_passages\} \\ \\ Question | Answer | Citation:}
    }
} \\ \\
For inputs containing images, we use the following prompt as the system prompt:\\ \\
\fbox{
    \parbox{0.98\textwidth}{
        \texttt{You are given a question style description, which describes the characteristics of a specific style of reading comprehension question. You are also provided with one or more captioned images, and possibly some additional text or table passages. In addition, you are also provided with modality requirements. Your task is to generate a question following the style specified in the question style description based on the input images, image captions and text or table passages. Also generate an answer to the question and citations of the images or passages the answer is based on.\\ \\ RULES\\1. The question you generate MUST be based on the provided images, image captions and passages (if provided). The subset of provided images and passages that the question is based on are the chosen source materials.\\2. The modality requirements constrain the images and passages you can choose as the source materials. For example, if the modality requirement is 1 image, 1 text, your question MUST be based on 1 image (and its caption) and 1 text passage exactly.\\3. The question should not be answerable if any of the chosen source materials are removed.\\4. Generate a natural sounding question that a reasonable human being might ask.\\5. Produce a response in the following format: <question> | <answer> | <citation>. The citations should be a reference to the images or passages chosen. You should images and passages by their number only\\6. You MUST use the image captions of the source images you have chosen explicitly in the question. For example, if the caption says "Roger Federer", then "Roger Federer" must be explicitly stated in the question somewhere.\\7. Closely follow the question style description. If you cannot do this, say None. \\ \\ \{style\_prompt\}}
    }
} \\ \\
\noindent Few shot examples, text and table passages and images are passed to GPT-4-Turbo as conversation turns via the chat completion API. Images are directly captioned, and non-image data is formatted according to the following template:\\ \\
 \fbox{
    \parbox{0.98\textwidth}{
        \texttt{Passages: \{enumerated\_passages\} \\ \\ Question | Answer | Citation:}
    }
}\\

\subsection{Question Style Prompts}

The question style prompts contain descriptions $v$ and are inserted into the question generation prompts at \texttt{\{style\_prompt\}}. Creating questions with new styles only involves writing new question style prompts along with corresponding few-shot examples. In this section, we provide the question style prompts used to generate our SMMQG Wikipedia dataset.

\subsubsection{Info Extraction}
\fbox{
    \parbox{0.98\textwidth}{\texttt{Question style: Information extraction question. Simple question that can be answered by extracting a fact from a single passage or image, if provided. Examples of such questions: Who is the founder of Microsoft? On what date did the Battle of Stalingrad begin? Which two actors won the Academy Awards for Best Actor and Best Supporting Actor in 2012?}
    }
} \\

\subsubsection{Compare Contrast}
\fbox{
    \parbox{0.98\textwidth}{\texttt{Question style: Compare and contrast question. Requires making comparisons based on information from one or more passages or images, if provided. The two subjects being contrasted must belong to the same category and be directly comparable - being about the same topic is not enough. Do not create questions about subjects that are not closely related and do not belong to the same category - if you cannot be sure, say None. Examples of such question: Compare and contrast the embryonic development process of humans and monkeys. Compare and contrast the careers of tennis players Roger Federer and Rafael Nadal. Format your answer to this question like this: explain what the relationship between the two subjects are and why they are similar. Next, identify several comparable traits and explain how the traits of the subjects are similar or different. Avoid simply summarising one subject after the other - make sure to interweave both subjects in your answer.}
    }
} \\

\subsubsection{Numerical}
\fbox{
    \parbox{0.98\textwidth}{\texttt{Question style: Maths question. Requires calculation based on numbers from passages to determine the answer. Calculation must be used to answer the question, simple extraction of numbers is not enough. If there are no numbers mentioned in the passages, say None. If there are no calculations possible, say None. Examples of such questions include: how many more trophies did Barcelona win between 2000 and 2010 then Atleti? What was the percentage change in the number of Covid cases in Italy between March 2020 and September 2020?}
    }
} \\

\subsubsection{Compound}
\fbox{
    \parbox{0.98\textwidth}{\texttt{Question style: Compound question. Question is composed of two thematically related subquestions connected by ``and". Examples of such question include: How many Grand Slams did Roger Federer win, and how many of those were at the US Open? Who is the first King of England, and what is the significance of the Crown Jewels to English royalty? What is a hurricane, and what weather event causes the most deaths in a typical year?}
    }
}

\onecolumn\section{QA Verification}
\label{sec:appendix_qa_verification}

For inputs containing text or tables only, the prompt is:\\ \\ 
\fbox{
    \parbox{0.98\textwidth}{\texttt{You are given a question, an answer to that question, and some supporting passages. You are also given a question style, some information about it and some examples of questions in that style. Your job is to assess whether the question and answer passes or fails, based on 2 criteria:\\ \\Criterion 1. The answer to the question can be inferred from the supporting passages provided.\\Criterion 2. The question closely matches the style of question specified by the question style.\\ \\If all 2 criteria are met, return Pass. If one or more conditions are not met, return Fail, along with the list of conditions that were not met.\\ \\ <few shot examples>\\ \\Question: \{question\} \\Answer: \{answer\}\\Question Style: \{style\_prompt\}\\Passages: \{enumerated\_passages\} \\ \\Assessment:}
    }
} \\ \\

\noindent For inputs containing images, the following prompt is used as the system prompt: \\ \\
\fbox{
    \parbox{0.98\textwidth}{\texttt{You are given a question, an answer to that question, and some supporting images and (optionally) passages. You are also given a question style, some information about it and some examples of questions in that style. Your job is to assess whether the question and answer passes or fails, based on 2 criteria:\\ \\ Criterion 1. The answer to the question can be inferred from the supporting images and (optionally) passages provided.\\Criterion 2. The question closely matches the style of question specified by the question style.\\ \\If all 2 criteria are met, return Pass. If one or more conditions are not met, return Fail, along with the list of conditions that were not met.}
    }
} \\ \\

\noindent Few shot examples, questions, answers, question styles, text and table passages and images are passed to GPT-4-Turbo as conversation turns via the chat completion API. Images are directly captioned, and non-image data is formatted according to the following template:\\ \\
 \fbox{
    \parbox{0.98\textwidth}{
        \texttt{Question: \{question\}\\Answer: \{answer\}\\Question Style: \{style\_prompt\}\\Passages: \{enumerated\_passages\}\\ \\ Assessment:}
    }
}

\onecolumn\section{Multi-hop Question Generation}
\label{sec:appendix_multi_hop_qg}

\begin{figure*}
    \centering
    \includegraphics[width=1.0\textwidth]{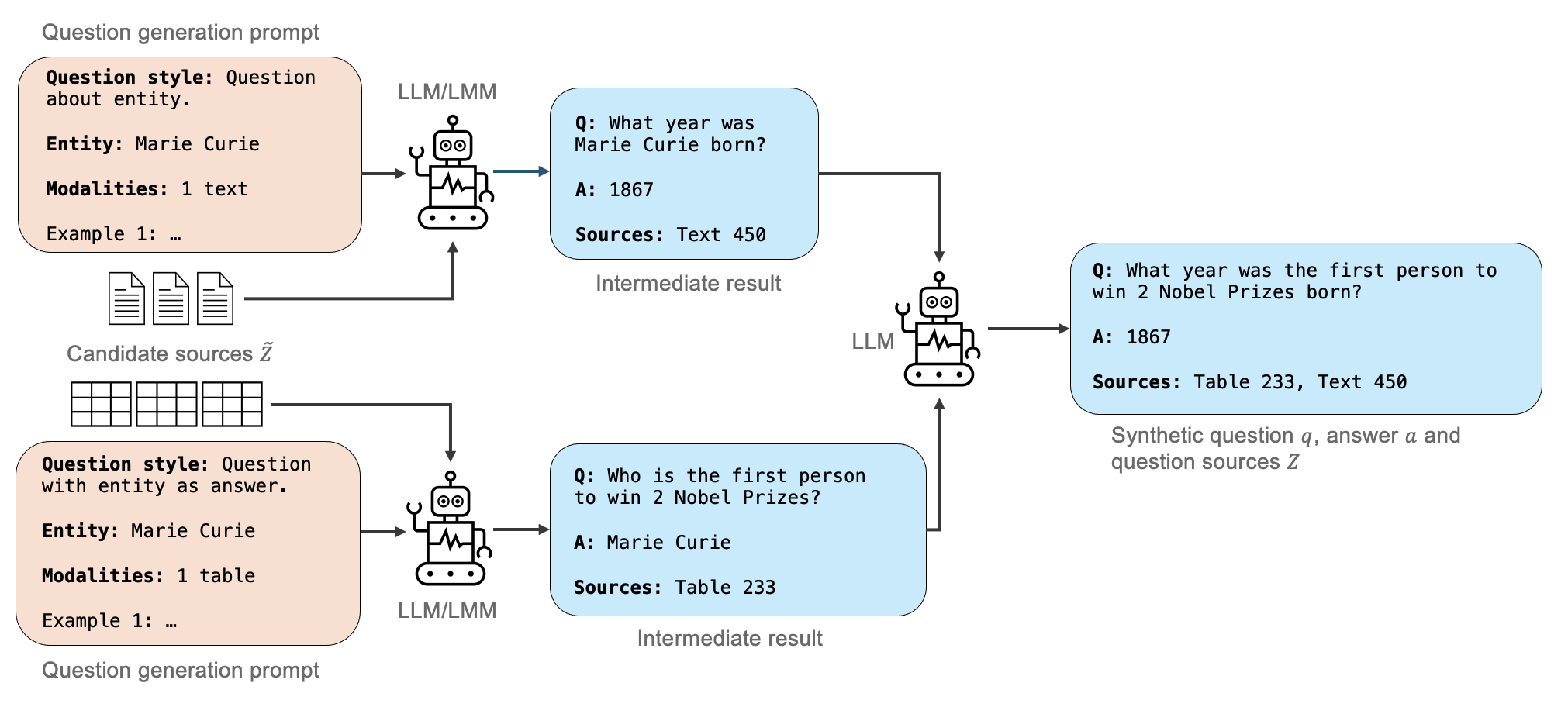}
    \caption{\textbf{Multi-hop Question Generation}. Instead of directly producing a question as in standard SMMQG, we first generate two intermediate questions. The first question is about the extracted entity, and the second has the extracted entity as the answer. These are then combined by an LLM or LMM to produce the final, multi-hop question.}
    \label{fig:multi_hop_diagram}
\end{figure*}

\subsection{Intermediate Question Generation Prompts}

The intermediate question generation steps for multi-hop question generation use the same question generation prompts as for standard question generation, but with specific question style prompts. Unlike before, we also insert the entity found in Step~2 into the question style prompt:

\subsubsection{Question About Entity Prompt}
\fbox{
    \parbox{0.98\textwidth}{\texttt{Question style: Information extraction question about an entity: \{entity\}. Simple question that can be answered by extracting a fact from a single passage or image about the provided entity. The question you generate MUST contain a direct and explicit reference to the entity. Avoid generating questions asking what/who is entity e.g. do not ask "Who is Roger Federer", but ask something about Roger Federer explicitly. Also avoid generating questions not based on hard facts e.g. "what is company X known for", "what is person Y famous for". If you cannot generate a question about this entity, say None. Examples of such entity and question combinations: Entity: Microsoft. Question: Who is the founder of Microsoft? Entity: Battle of Stalingrad. Question: On what date did the Battle of Stalingrad begin? Entity: Academy Awards 2012. Question: Which actor won the Academy Award for Best Actor in 2012?}
    }
}

\subsubsection{Question With Entity As Answer Prompt}
\fbox{
    \parbox{0.98\textwidth}{\texttt{Question style: Information extraction question where the answer is the provided entity: \{entity\}. Simple fact extraction question where the answer is the entity provided. The answer you give must be the entity itself, nothing more. If you cannot generate a question with the entity as the answer, say None. Try your best to create a question where the entity is likely the unique answer - this will require you to think about whether there might be other entities that can be used to answer the question. If such entities exist, ask a different question. Examples of such entity and question combinations: Entity: Microsoft. Question: Which company developed the Windows OS? Answer: Microsoft. Entity: Stalingrad. Question: What was the name of the city of Volgograd during the Second World War? Answer: Stalingrad.}
    }
}

\subsection{Combination Prompt}

The combination prompt is used to combine the two intermediate questions into a multi-hop question via an LLM. For inputs containing text or tables only:\\ \\
\fbox{
    \parbox{0.98\textwidth}{\texttt{You are given an entity and two questions, answers and passages. The first question is a question about the entity, and the second question is a question with an answer that is the entity. Your task is to combine the questions into a single, multi-hop question, and combine the answers in order to create an answer to the multi-hop question. A multi-hop question is one that requires answering an implicit sub-question before the full question can be answered. For example, the question "Who is the wife of the 40th president of the US" is a multi-hop question, because it requires first answering "Who is the 40th president of the US?" and using that answer to answer the full question. You can always construct a multi-hop questions from two questions if one of the questions contains a direct reference to an entity that is also the answer to the second question. The procedure simply involves replacing the reference to the common entity in the first question with the second question.\\ \\RULES\\1. Combine the questions into a multi-hop question, if possible.\\2. Combine the answers into a single, answer that answers the multi-hop question step-by-step.\\3. If the two provided questions do not fulfill the criteria for constructing a multi-hop question (one question contains an explicit reference to the entity, the other has as its answer the entity), return None.\\4. Try and phrase the multi-hop question as naturally as possible without altering the validity of the answer, and make sure that the generated multi-hop question makes sense. If this is not possible and the question is deemed awkward, return None.\\5. Return your multi-hop question and answer in the format: <multi-hop question | answer>.\\6. If the second question can be answered by the first passage, or the first question can be answered by the second passage, return None.\\7. If there no need to resolve the implicit question in order to resolve the full question, return None.\\8. Check that the questions can be combined to form a meaningful question. Sometimes the individual questions are valid but cannot be combined. If the combined question is not valid, say None.\\9. If you cannot fulfill any of the rules above, return None. It is preferable to return None than to break any of the rules.\\ \\ <few shot examples> \\ \\ Passages: \{enumerated\_passages\} \\ \\ Question 1: \{question\_1\} \\ Question 2: \{question\_2\} \\ Answer 1: \{answer\_1\} \\ Answer 2: \{answer\_2\} \\ Entity: \{entity\} \\ \\ Multi-hop Question | Answer:}
    }
} \\ \\

\noindent For inputs containing images, the following prompt is used as the system prompt: \\ \\
\fbox{
    \parbox{0.98\textwidth}{\texttt{You are given an entity and two questions, two answers, some images and possibly some passages. The first question is a question about the entity, and should contain an explict reference to the entity in it. The second question is a question with an answer that is the entity. Your task is to combine the questions into a single, multi-hop question, and combine the answers in order to create an answer to the multi-hop question. A multi-hop question is one that requires answering an implicit sub-question before the full question can be answered. For example, the question "Who is the wife of the 40th president of the US" is a multi-hop question, because it requires first answering "Who is the 40th president of the US?" and using that answer to answer the full question. You can always construct a multi-hop questions from two questions if one of the questions contains a direct reference to an entity that is also the answer to the second question. The procedure simply involves replacing the reference to the common entity in the first question with the second question.\\ \\ RULES\\1. Combine the questions into a multi-hop question, if possible.\\2. Combine the answers into a single, answer that answers the multi-hop question step-by-step.\\3. If the two provided questions do not fulfill the criteria for constructing a multi-hop question (one question contains an explicit reference to the entity, the other has as its answer the entity), return None.\\4. If the answer to the first question (the one that should be containing an explicit reference to the entity) also has an answer that is the entity, return None.\\5. If the entity provided is not the entity that bridges the two questions (i.e. is referenced in one and is the answer to the other), return None.\\6. Try and phrase the multi-hop question as naturally as possible without altering the validity of the answer, and make sure that the generated multi-hop question makes sense. If this is not possible and the question is deemed awkward, return None.\\7. Return your multi-hop question and answer in the format: <multi-hop question | answer>.\\8. The passages/images associated with the first question are labelled/captions Passage 1/Image 1. The passages/images associated with the second question are labelled/captions Passage 2/Image 2. If Passage 1/Image 1 can be used to answer the second question, or vice versa, return None.\\9. If you cannot fulfill any of the rules above, return None. It is preferable to return None than to break any of the rules.}
    }
} \\

\noindent Few shot examples, questions, answers, passages, entities and images are passed to GPT-4-Turbo as conversation turns via the chat completion API. Images are directly captioned, and non-image data is formatted according to the following template:\\ \\
\fbox{
    \parbox{0.98\textwidth}{\texttt{Passages: \{enumerated\_passages\} \\ \\ Question 1: \{question\_1\} \\ Question 2: \{question\_2\} \\ Answer 1: \{answer\_1\} \\ Answer 2: \{answer\_2\} \\ Entity: \{entity\} \\ \\ Multi-hop Question | Answer:}
    }
}

\section{GPT-4-Turbo Judge}
\label{sec:appendix_judge}

For inputs containing text or tables only:\\ \\
\fbox{
    \parbox{0.98\textwidth}{\texttt{You are a fair and unbiased judge. You have been given a question, a model answer to that question and some source passages. The source passages should contain all the information required to answer the question. You are then provided with a candidate answer to the question. Your objective is to score the candidate answer to the question. Use the sources and the model answer to better understand what the truth is and assign the candidate answer a score. Return an explanation followed by the score. Separate the explanation from the score with ``Score:"\\ \\IMPORTANT: the model answer should only act as a guide. It is possible for the new answer to be different from the model answer but still be correct. You must think about the question and look at the source passages closely.\\ \\- Provide a score of 0 if the candidate answer is incorrect.\\- Provide a score of 1 if the candidate answer is somewhat correct, but is missing something important or contains some minor inaccuracies.\\- Provide a score of 2 if the candidate answer is correct.\\ \\<few shot examples>\\ \\Question: \{question\}\\Candidate Answer: \{candidate\_answer\}\\Model Answer: \{model\_answer\}\\Passages: \{passages\}\\ \\Explanation and Score:}
    }
} \\ \\

\noindent For inputs containing images, the following prompt is used as the system prompt:\\ \\
\fbox{
    \parbox{0.98\textwidth}{\texttt{You are a fair and unbiased judge. You have been given a question, a model answer to that question and one or more images and possibly some passages. The images and passages together form the sources, and this should contain all the information required to answer the question. You are then provided with a candidate answer to the question. Your objective is to score the candidate answer. Use the sources and the model answer to better understand what the truth is and assign the candidate answer a score. Return an explanation followed by the score. Separate the explanation from the score with ``Score:".\\ \\IMPORTANT: the model answer should only act as a guide. It is possible for the candidate answer to be different from the model answer but still be correct. You must think about the question and look at the images and passages closely.\\ \\- Provide a score of 0 if the candidate answer is incorrect.\\- Provide a score of 1 if the candidate answer is somewhat correct, but is missing something important or contains some minor inaccuracies.\\- Provide a score of 2 if the candidate answer is correct.\\ \\}
    }
}\\ \\

\noindent Few shot examples, questions, passages, answers and images are passed to GPT-4-Turbo as conversation turns via the chat completion API. Images are directly captioned, and non-image data is formatted according to the following template:\\ \\
\fbox{
    \parbox{0.98\textwidth}{\texttt{Question: \{question\}\\Candidate Answer: \{candidate\_answer\}\\Model Answer: \{model\_answer\}\\Passages: \{passages\}\\ \\Explanation and Score:}
    }
}\\

\section{Evaluation Results with Additional Metrics}
\label{sec:appendix_additional_judge}

In addition to GPT-4 scores, we also report evaluation results computed using GPT-3.5-Turbo-judge, BERTScore and ROUGE scores. For GPT-3.5-Turbo, we ask the judge to compare the model answer (reference) against the predicted answer given the question and to score the answer as correct, incorrect or partially correct. Our GPT-3.5.-Turbo judge prompt is as follows:\\ \\
\fbox{
    \parbox{0.98\textwidth}{\texttt{You are a fair judge. You are provided with a question, a reference answer and a prediction. Decide whether the prediction is correct based on the reference answer on a scale of 0 to 2. If the prediction is incorrect, return a 0. If the prediction is partially correct, return a 1. If the prediction is correct, return a 2. Do not use your own knowledge. Use only the reference.\\ \\Example 1\\Question: Who was the F1 champion in 2019?\\Reference: Lewis Hamilton\\Prediction: Seb Vettel\\Score: 0\\ \\Example 2\\Question: Explain the purpose of dropout in neural networks.\\Reference: Dropout is used for regularisation. It helps prevent overfitting.\\Prediction: Reduction in overfitting via regularisation.\\Score: 2\\ \\Example 3\\Question: Compare and contrast the careers of Federer and Nadal.\\Reference: Federer has won 20 grand slams, and Nadal has won 22. Federer retired in 2021, whereas Nadal currently still plays tennis.\\Prediction: Federer won 20 grand slams and Nadal won 22.\\Score: 1\\ \\Input\\Question: \{question\}\\Reference: \{reference\}\\Prediction: \{prediction\}\\Score: }
    }
}\\ \\

\noindent As GPT-3.5-Turbo lacks multimodal capabilities, we do not provide sources for decision-making - the judge assesses model performance based only on the model answer. For BERTScore, we use \texttt{microsoft/deberta-xlarge-mnli} as the BERTScore model and treat the model answer as the reference and the predicted answer as the candidate. For ROUGE evaluation, we compute ROUGE-1 (unigram overlap), again between the model answer and the predicted answer.

We report aggregate results for each model in Table \ref{tab:additional_judges} and report Kendall's tau on the evaluation ranked lists against GPT-4-Turbo judge in Table \ref{tab:additional_judges_kendall}

\begin{table}[h]
    \small
    \centering
    \begin{tabular}{p{0.23\linewidth} | c c c c}
    \hline
     & \textbf{GPT-4-Turbo Judge} & \textbf{GPT-3.5-Turbo Judge} & \textbf{BERTScore} & \textbf{ROUGE} \\ \hline
    Vicuna-7b + LLaVA-7b & 45.3 & 56.9 & 61.7 & 53.0 \\
    Vicuna-13b + LLaVA-13b & 61.3 & 68.6 & 69.0 & 61.3 \\
    Qwen-Chat + Qwen-VL-Chat & 61.1 & 69.2 & 66.8 & 66.2 \\ \hline
    Gemini Pro 1.0 & 73.9 & 77.1 & \textbf{70.2} & 63.3 \\
    Claude 3 Haiku & 62.7 & 71.5 & 62.9 & 71.9 \\
    Claude 3 Sonnet & 74.1 & 80.3 & 66.2 & 70.9 \\
    Claude 3 Opus & 86.5 & 82.7 & 68.2 & 72.0 \\ \hline
    GPT-4-Turbo & \textbf{93.4} & \textbf{87.8} & 69.8 & \textbf{77.9} \\ \hline
    \end{tabular}
    \caption{Evaluation results with other evaluation metrics. We report results aggregated over all question styles and question modalities.}
    \label{tab:additional_judges}
\end{table}

\begin{table}[h]
    \small
    \centering
    \begin{tabular}{p{0.12\linewidth} | c c c}
    \hline
     &  \textbf{GPT-3.5-Turbo Judge} & \textbf{BERTScore} & \textbf{ROUGE} \\ \hline
    $\tau$ & 0.93 & 0.50 & 0.72 \\
    $p$-value & <0.001* & 0.11 & 0.014* \\ \hline
    \end{tabular}
    \caption{Kendall's tau and its associated $p$-value on the ranked lists of evaluation results against GPT-4-Turbo judge. We find that GPT-3.5-Turbo judge and ROUGE correlate well with GPT-4-Turbo judge.}
    \label{tab:additional_judges_kendall}
\end{table}

\section{Analysis of Question Style Overlap}
\label{sec:appendix_q_overlap}

In order to better understand how distinct our pre-defined question styles are, we analyse the level of question style overlap in our SMMQG-generated dataset. We use GPT-4-Turbo as a question style binary classifier, pass each QA pair along with descriptions of the question styles to it and ask whether the question belongs to a given style. We define questions with overlap as those that elicit a positive response from the classifier for two or more styles, and report the percentage of questions in our dataset that contain overlap in Table \ref{tab:q_overlap}, breaking down results by the ground-truth style:

\begin{table}[h]
    \small
    \centering
    \begin{tabular}{p{0.15\linewidth} | c c c c c}
    \hline
     &  Info Extraction & Compare Contrast & Numerical & Compound & Multi-hop \\ \hline
    \textbf{\% Overlap} & 4.3 & 0.0 & 4.9 & 2.8 & 20.0 \\ \hline
    \end{tabular}
    \caption{Percentage of questions with overlap by style. We define questions with overlap as those that are classified by GPT-4-Turbo as belonging to more than one question style. The only question style with significant overlap is the \textit{Multi-hop} question style.}
    \label{tab:q_overlap}
\end{table}

We see that overlap is insignificant for all but the multi-hop questions. Upon closer inspection, we notice that the multi-hop style commonly overlaps with the numerical and info extraction styles. Our explanation is that multi-hop reasoning inherently requires use of other reasoning skills because these other skills are needed to solve the individual subparts of multi-hop questions.

\section{Model Details and Licences}
\label{sec:appendix_model_details}

\begin{table}[h]
    \small
    \centering
    \begin{tabular}{| p{0.23\linewidth} | p{0.5\linewidth} | p{0.15\linewidth}|}
    \hline
    \textbf{Name} & \textbf{Hugging Face ID} & \textbf{Licence}\\ \hline
    E5-Large & \texttt{intfloat/e5-large-v2} & mit \\ \hline
    OpenCLIP & \texttt{laion/CLIP-ViT-H-14-laion2B-s32B-b79K} & mit \\
    \hline
    \end{tabular}
    \caption{Details for retriever models.}
    \label{tab:model_details_1}
\end{table}

\begin{table}[h]
    \small
    \centering
    \begin{tabular}{| p{0.23\linewidth} | p{0.5\linewidth} |p{0.15\linewidth} |}
    \hline
    \textbf{Name} & \textbf{Details} & \textbf{Licence} \\ \hline
    LLaVA-13b (Verbalisation) & \texttt{liuhaotian/llava-llama-2-13b-chat-lightning-preview} & LLAMA 2 Community \\ \hline
    LLaVA-v1.5-7b & \texttt{liuhaotian/llava-v1.5-7b} & LLAMA 2 Community \\ \hline
    LLaVA-v1.5-13b & \texttt{liuhaotian/llava-v1.5-13b} & LLAMA 2 Community \\ \hline
    Vicuna-7b-v1.5 & \texttt{lmsys/vicuna-7b-v1.5} & LLAMA 2 Community \\ \hline
    Vicuna-13b-v1.5 & \texttt{lmsys/vicuna-13b-v1.5} & LLAMA 2 Community \\ \hline
    Qwen-Chat & \texttt{Qwen/Qwen-7B-Chat} & Tongyi Qianwen \\ \hline
    Qwen-Chat-VL & \texttt{Qwen/Qwen-VL-Chat} & Tongyi Qianwen \\ \hline
    \end{tabular}
    \caption{Details for open-source LLMs and LMMs.}
    \label{tab:model_details_2}
\end{table}

\begin{table}[h]
    \small
    \centering
    \begin{tabular}{| p{0.25\linewidth} | p{0.5\linewidth} |}
    \hline
    \textbf{Name} & \textbf{Details} \\ \hline
    GPT-4-Turbo & \texttt{gpt-4-1106-vision-preview} \\ \hline
    Gemini Pro 1.0 & \texttt{gemini-1.0-pro-vision-001} \\ \hline
    Claude 3 Haiku & \texttt{anthropic.claude-3-haiku-20240307-v1:0} \\ \hline
    Claude 3 Sonnet & \texttt{anthropic.claude-3-sonnet-20240229-v1:0} \\ \hline
    Claude 3 Opus & \texttt{anthropic.claude-3-opus-20240229-v1:0} \\ \hline
    \end{tabular}
    \caption{Details for API-based models.}
    \label{tab:model_details_3}
\end{table}

All model evaluation experiments were done using \textbf{greedy decoding}. All inference experiments with open-source models were done using 1 A100 GPU. We document model details and licenses in Tables \ref{tab:model_details_1}, \ref{tab:model_details_2} and \ref{tab:model_details_3}.

\section{Additional Dataset}\label{sec:appendix_bio_dataset}

\begin{table*}[hbt!]
    \small
    \centering
    \begin{tabular}{p{0.15\linewidth} | c c c c c}
    \hline
      & Q. Fluency & Q. Style Faithfulness & Source Relevance & Answerability & A. Correctness \\ \hline
     SMMQG & 4.47 & 94.7 & 86.4 & 91.2 & 91.2 \\ \hline
    \end{tabular}
    \caption{\textbf{Human study results on our additional dataset}. These results show that our biology SMMQG dataset quality is high as assessed by crowdworkers.}
    \label{tab:additional_human_study}
\end{table*}

In addition to our main Wikipedia-derived SMMQG dataset, we also generate a separate QA dataset on a college-level biology textbook \citep{fowler2013concepts} containing text and image sources.\footnote{\texttt{https://dept.clcillinois.edu/biodv/PrinciplesOfBiology.pdf}} We use the exact same SMMQG parameters and prompts as for the original dataset to generate this dataset. Our additional dataset contains 324 questions derived from 862 individual sources. 

We conducted an additional human study to assess the quality of this dataset and thereby demonstrate that SMMQG generalises to other, more domain-specific sources than Wikipedia. We sampled 150 questions from our new dataset and asked crowdworkers to assess their quality, repeating the process described in Section \ref{sec:human_study}. We report our findings in Table \ref{tab:additional_human_study}. We find that the quality of our new dataset remains high despite the use of a more domain specific dataset. This demonstrates that SMMQG generalises well beyond Wikipedia.

\section{Concurrence Study}
\label{sec:appendix_concur_study}

\subsection{Evaluation Results and Discussion}

\begin{table}[h]
    \small
    \centering
    \begin{tabular}{p{0.1\linewidth} p{0.08\linewidth} | c c }
    \hline
     & & MMQA & SMMQG  \\ \hline
     BM25 & top-5 & 36.3 & \textbf{42.8} \\
        & top-10 & 42.1 & \textbf{45.0} \\
     E5 & top-5 & \textbf{54.3} & 53.5 \\
        & top-10 & \textbf{60.1} & 56.7 \\
     OpenCLIP & top-5 & 38.3 & \textbf{40.0} \\
        & top-10 & 43.7 & \textbf{45.3} \\ \hline
    \end{tabular}
    \caption{\textbf{Retrieval evaluation results} for our concurrence experiments in Section \ref{sec:concurrence}.}
    \label{tab:mmqa_retrieval_comparison}
\end{table}

\begin{table}[h]
    \small
    \centering
    \begin{tabular}{p{0.25\linewidth} | c c}
    \hline
     & MMQA & SMMQG  \\ \hline
     Vicuna-7b + LLaVA-7b & 55.0 & \textbf{68.0} \\
     Vicuna-13b + LLaVA-13b & 61.0 & \textbf{75.0}  \\
     Qwen-Chat + Qwen-VL-Chat & 65.7 & \textbf{78.5}  \\
     Gemini Pro 1.0 & 80.7 & \textbf{89.6} \\
     Claude 3 Haiku & 73.9 & \textbf{75.9}  \\
     Claude 3 Sonnet & \textbf{87.3} & 86.9  \\
     Claude 3 Opus & \textbf{93.0} & 92.9 \\
     GPT-4-Turbo & 96.4 & \textbf{97.4} \\     
    \hline
    \end{tabular}
    \caption{\textbf{QA evaluation results} for our concurrence experiments in Section \ref{sec:concurrence}.}
    \label{tab:mmqa_qa_comparison}
\end{table}

Tables \ref{tab:mmqa_retrieval_comparison} and \ref{tab:mmqa_qa_comparison} contain the SMMQG and MMQA evaluation results used in our concurrence study in Section \ref{sec:concurrence}. We see that information extraction and multi-hop SMMQG questions are in general easier to answer than their MMQA counterparts. One hypothesis is that this results from MMQA questions being less fluent and less answerable, as shown in Section \ref{sec:human_study}. To investigate this, we remove all samples with unanswerable questions and with question fluencies of less than 4 from the datasets used in Section \ref{sec:concurrence}, and replace them with new samples that we manually validate. We re-run evaluation on this new dataset, but find that MMQA questions remain more challenging than SMMQG questions. 

We conclude that, on comparable styles, our SMMQG questions are generally easier to answer than their MMQA counterparts. We hypothesise that this is the result of the question generation model picking the most ``obvious'' questions to ask, whereas crowdworkers may attempt to produce more novel questions. Nonetheless, we note that question style and modality have far more influence on the difficulty of questions, as evidenced by our results in Section \ref{sec:eval_experiments}, and that our SMMQG dataset holds similar discriminative power to MMQA despite being easier, as evidenced by our concurrence experiment results. Finally, we suggest that question difficulty may be increased via prompting through question style prompt $v$; we leave exploration of this to future work.

\section{Human Study}
\label{sec:appendix_human_study}

\subsection{Additional Results}

We present additional human study results breaking down performance by question style in Table \ref{tab:human_study_additional}. We find that SMMQG-question quality is maintained across question styles.

\begin{table*}[hbt!]
    \small
    \centering
    \begin{tabular}{p{0.15\linewidth} | c c c c c}
    \hline
      & Q. Fluency & Q. Style Faithfulness & Source Relevance & Answerability & A. Correctness \\ \hline
     Info Extraction & 4.73 & 100.0 & 100.0 & 96.7 & 95.0 \\
     Single Modality & 3.77 & 100 & 86.7 & 86.7 & 80.0 \\ \hline
     \centering $\Delta$ & \textbf{+0.96} & +0.0 & \textbf{+13.3} & +10.0 & \textbf{+15.0} \\ 
     \centering $p$-value & <0.001* & 1.0 & 0.006* & 0.09 & 0.025* \\ \hline
     \noalign{\vskip 3mm}    
     \hline
     Multi-hop & 4.47 & 98.3 & 91.7 & 100.0 & 95.0 \\
     Compose & 3.57 & 93.3 & 85.0 & 85.0 & 80.0 \\ \hline 
     \centering $\Delta$ & \textbf{+0.90} & +5.0 & +6.7 & \textbf{+15.0} & \textbf{+15.0} \\ 
     \centering $p$-value & <0.001* & 0.44 & 0.39 & 0.003* & 0.025* \\ \hline
    \noalign{\vskip 3mm}    
    \hline
     Compare Contrast & 4.52 & 98.3 & 88.3 & 88.3 & 88.3 \\
     Numerical & 4.50 & 100.0 & 88.3 & 91.7 & 91.7 \\
     Compound & 4.43 & 95.0 & 96.7 & 96.7 & 93.3 \\ \hline
    \end{tabular}
    \caption{\textbf{Human study results by question style}. In the top and middle subtables, we compare \textit{single modality} and \textit{compose} questions from MMQA with SMMQG info extraction and multi-hop questions. In the bottom subtable, we report results for the remaining SMMQG styles. We denote statistically significant differences in \textbf{bold} and $p$-values with $p \leq 0.05$ using *.}
    \label{tab:human_study_additional}
\end{table*}

\subsection{Additional Details}

We recruited 25 crowdworkers for our human study via crowdsourcing platform Prolific (\color{blue}\texttt{www.prolific.com})\color{black}. Each crowdworker was given a total of 90 minutes to read through the instructions and evaluate 20 samples, for which they were paid 17 GBP (with the added possibility of a bonus). We screened crowdworkers and selected only those (1) located in the US, UK, Ireland, Australia, New Zealand and Canada with English as their primary language and (2) possessing at least a Bachelor's degree.\\

\noindent We conducted our human study via Google Forms. Each crowdworker was sent a link to a Google Form containing 20 samples and 3 attention checks, along with a further link to the instructions. The instructions were provided via a Google Doc that contained an in-depth explanation of the task, the evaluation metrics, the bonus structure, as well as three fully-worked examples.\\

\noindent Of the 20 samples, 3 were always test samples of varying difficulties. These were included in order to assess the quality of responses from each crowdworker. Crowdworkers that correctly labelled all 3 test samples were awarded a bonus of 8 GBP. We manually reviewed the responses of crowdworkers who incorrectly labelled 2 or more test samples and made the appropriate corrections (at no point during this review process were we made aware of the source of any given question). Of the 25 crowdworkers, 8 scored 3/3 on the test questions, 13 scored 2/3 and 4 scored 1/3 or below.\\

\noindent The remaining 17 samples in each form were drawn without replacement from a pool of SMMQG and MMQA samples. This pool was in turn composed of 120 MMQA and 300 SMMQG samples, with 60 samples from each of the two MMQA and five SMMQG styles present. These samples were drawn randomly without replacement from the full datasets. The final crowdworker was shown only 12 rather than 17 samples in order to maintain an even distribution over styles. \\ \\

\begin{figure*}[hbt!]
    \centering
    \includegraphics[width=0.6\textwidth]{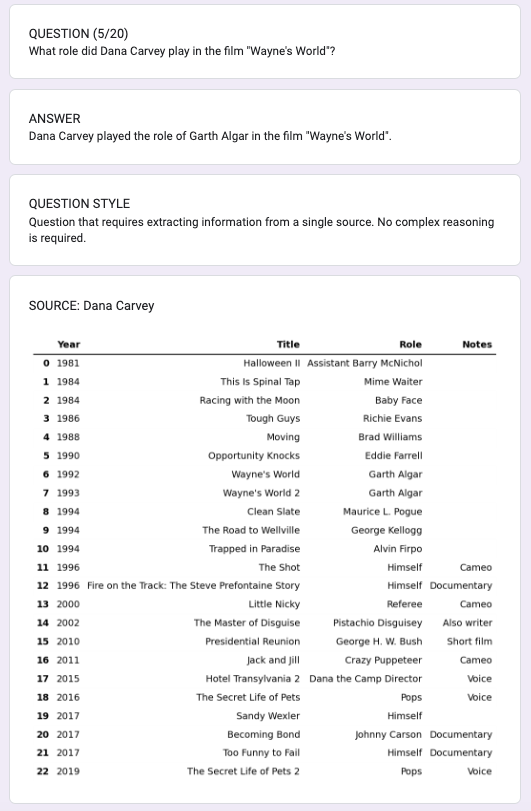}
    \caption{An example section from a Google Form shown to one of the crowdworkers. This section contains the question, answer, question style and source.}
    \label{fig:human_study_form_1}
\end{figure*}

\begin{figure*}[hbt!]
    \centering
    \includegraphics[width=0.6\textwidth]{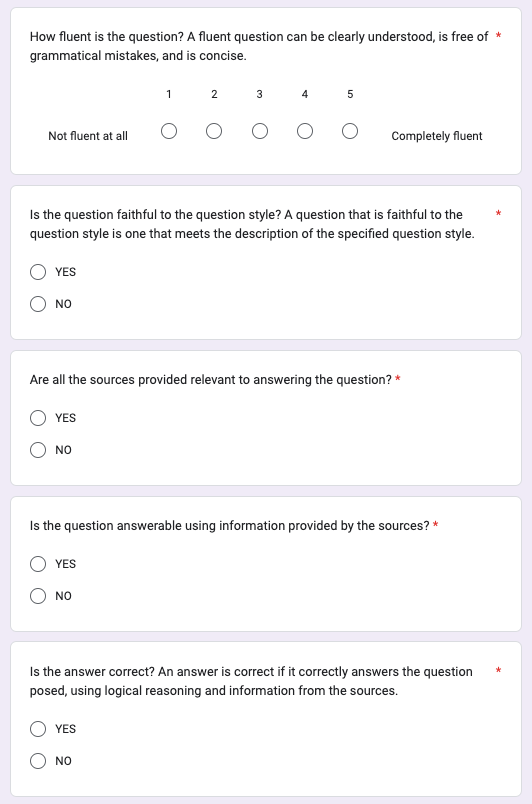}
    \caption{An example section from a Google Form shown to one of the crowdworkers. This section contains the response section.}
    \label{fig:human_study_form_2}
\end{figure*}

\clearpage

\noindent Crowdworkers were given these instructions via a Google Doc: \\ \\
\fbox{
    \parbox{0.98\textwidth}{\texttt{In this study, we compare various methods for creating questions and answers for reading comprehension tasks. We have used these methods to create some questions and answers. We would like your help deciding how good these questions and answers actually are. You will be given some questions and answers. Each question and answer pair is connected to one or two sources of information, which may take the form of text, tables or images, and a question style. Your job is to assess the question and answer along 5 criteria. There are 20 samples in total. We recommend spending on average between 3-4 minutes per sample. Some samples should require less time than this, and others more. Please read the following instructions carefully and make sure you understand the task before proceeding. We have factored in 15 minutes for you to read these instructions. The entire task, including instructions-reading, should take no more than 90 minutes to complete. Your completion code will be provided at the end of the Google form.\\ \\ There are 5 different criteria to assess a sample on. \\ \\ <continued on next page>}
    }
} \\
    
\fbox{
    \parbox{0.92\textwidth}{\texttt{\textbf{Question Fluency:} Here we assess how fluent the question is. A completely fluent question is one that is free of grammatical mistakes, is phrased well and is concise. Your task is to decide how fluent the question is on a scale of 1 (incomprehensible) to 5 (fluent). Use your best judgment, but a question that is free of grammatical mistakes and is phrased well should be rated a 5, and a question that is incomprehensible should be rated a 1.\\ \\ \textbf{Question Style Faithfulness:} The questions were created based on a question style, which describes the kind of question that should be created. The question style and its description is provided along with the question and answer. Your task is to decide if the question adheres to the question style specified.\\ \\ \textbf{Source Relevance:} The questions were created based on one or two sources, and should be answerable using information provided in these sources. The sources may be a mix of text, table or image sources. Your task is to decide if all the sources contain information that is relevant to answering the question. If any source contains only irrelevant or tangentially relevant information, you should answer NO. If all the sources are all relevant and provide useful information, you should answer YES. If a source contains information about the correct topic but nothing useful can be extracted from it, you should still answer NO. If there are two sources, both of which contain the same useful information, you should answer YES.\\ \\ \textbf{Answerability:} The questions were created based on one or two sources, and should be answerable using information provided in these sources. Your task is to decide if the sources provide enough information to produce a reasonable answer to the question. Do not use your own knowledge of the topic to decide whether or not the sources provide enough information. You should only look at the sources and decide if there is adequate information present to at least superficially answer the question. Reasoning from information contained in the sources is allowed, so long as all the facts can be derived from the sources.\\ \\ \textbf{Answer Correctness:} The answer to the question is created along with the question. Here we assess whether the answer to the question is correct given the sources. Your task is to decide whether or not the answer to the question is correct using information from the sources. The answer does not need to be perfect for it to be correct. You should mark it as correct so long as it provides a reasonable answer to the question. Do not use your own knowledge of the topic to assess the answer. You should put yourself in the shoes of a reasonably intelligent person who has no knowledge of the topic at hand, and use only information contained in the sources, even if it is outdated or wrong. Reasoning from information contained in the sources is allowed, so long as all the facts can be derived from the sources.\\ \\ <Working Examples>}
    }
}\\ \\

\end{document}